\documentclass[letterpaper]{article} % DO NOT CHANGE THIS
\usepackage{aaai2027}  % DO NOT CHANGE THIS
\usepackage[hyphens]{url}  % DO NOT CHANGE THIS
\usepackage{graphicx} % DO NOT CHANGE THIS
\usepackage{amsmath}
\usepackage{amssymb}
\usepackage{natbib}  % DO NOT CHANGE THIS AND DO NOT ADD ANY OPTIONS TO IT
\usepackage{caption} % DO NOT CHANGE THIS AND DO NOT ADD ANY OPTIONS TO IT
\usepackage{multirow}
\usepackage{algorithm}
\usepackage{algorithmic}
\usepackage{newfloat}
\usepackage{listings}
\usepackage{cuted}

\DeclareCaptionStyle{ruled}{labelfont=normalfont,labelsep=colon,strut=off} % DO NOT CHANGE THIS
\floatstyle{ruled}
\newfloat{listing}{tb}{lst}{}
\floatname{listing}{Listing}

\usepackage{booktabs}

\usepackage{pifont}
\title{DynamicWAM: Dual-Path Motion Conditioning for World--Action Models\\
in Dynamic Manipulation}

\usepackage{pifont}

\title{DynamicWAM: Dual-Path Motion Conditioning for World--Action Models\\
in Dynamic Manipulation}

\newcommand{\equalmark}{\textsuperscript{*}}
\newcommand{\advisormark}{\textsuperscript{\textdagger}}
\newcommand{\corrmark}{\textsuperscript{\ding{41}}}
\newcommand{\affmark}[1]{\textsuperscript{\rm #1}}

\author{
Yunfan Lou\equalmark\advisormark\affmark{1,4},
Hewen Gao\equalmark\affmark{1,3},
Xiyu Zhu\equalmark\affmark{1,2},
Zhuoran Qiao\affmark{1,3},
Xuan Han\affmark{1,2},
Yifan Yang\affmark{1},\\
Yifan Ye\affmark{2},
Boxian Yao\affmark{1,5},
Zhibo Pang\corrmark\affmark{1,2,3}
}

\affiliations{
\affmark{1}PKU-PI Lab
\quad
\affmark{2}Peking University
\quad
\affmark{3}ePyBot Intelligence
\quad
\affmark{4}National University of Singapore
\quad
\affmark{5}Tsinghua University\\[5pt]
\equalmark Equal contribution
\quad
\advisormark Project advisor
\quad
\corrmark Corresponding author
}

\begin{document}

\maketitle

\begin{abstract}
Dynamic manipulation requires robots to infer target motion and respond promptly, yet existing World--Action Models (WAMs) typically condition only on the current frame and execute large backbones synchronously, limiting motion awareness and responsive control in dynamic scenes. We propose DynamicWAM, a compact WAM for dynamic object manipulation with dual-path motion conditioning. DynamicWAM introduces history-flow conditioning, encoding temporally aligned optical-flow frames alongside the current observation through a frozen pretrained video VAE to preserve spatial motion structure, while injecting kinematic descriptors of displacement, duration, velocity, and acceleration into the action expert to provide motion magnitude and timing. The two complementary paths are fused through joint world--action attention. A distilled compact backbone and real-time chunking (RTC)-based asynchronous execution further enable responsive control. On DOMINO, DynamicWAM achieves a 38.2\% success rate and a 53.2 manipulation score, outperforming all evaluated baselines. Across 12 real-world tasks spanning linear, circular, and compound target motion, it achieves a 46.7\% average success rate, exceeding the strongest baseline by 22.9 percentage points. Project page: \url{https://dynamicwam.github.io/}.
\end{abstract}

\section{Introduction}

Robotic manipulation is increasingly moving beyond static tabletop settings toward environments where targets continue to move during inference and action execution~\cite{zitkovich2023rt}. This setting exposes two limitations of action-chunking policies: a single observation may not reveal the target's motion state, while the target may move during prediction and execution. Successful manipulation therefore requires both motion-aware future prediction and responsive control.

\begin{figure}[t]
    \centering
    \includegraphics[width=1\columnwidth]{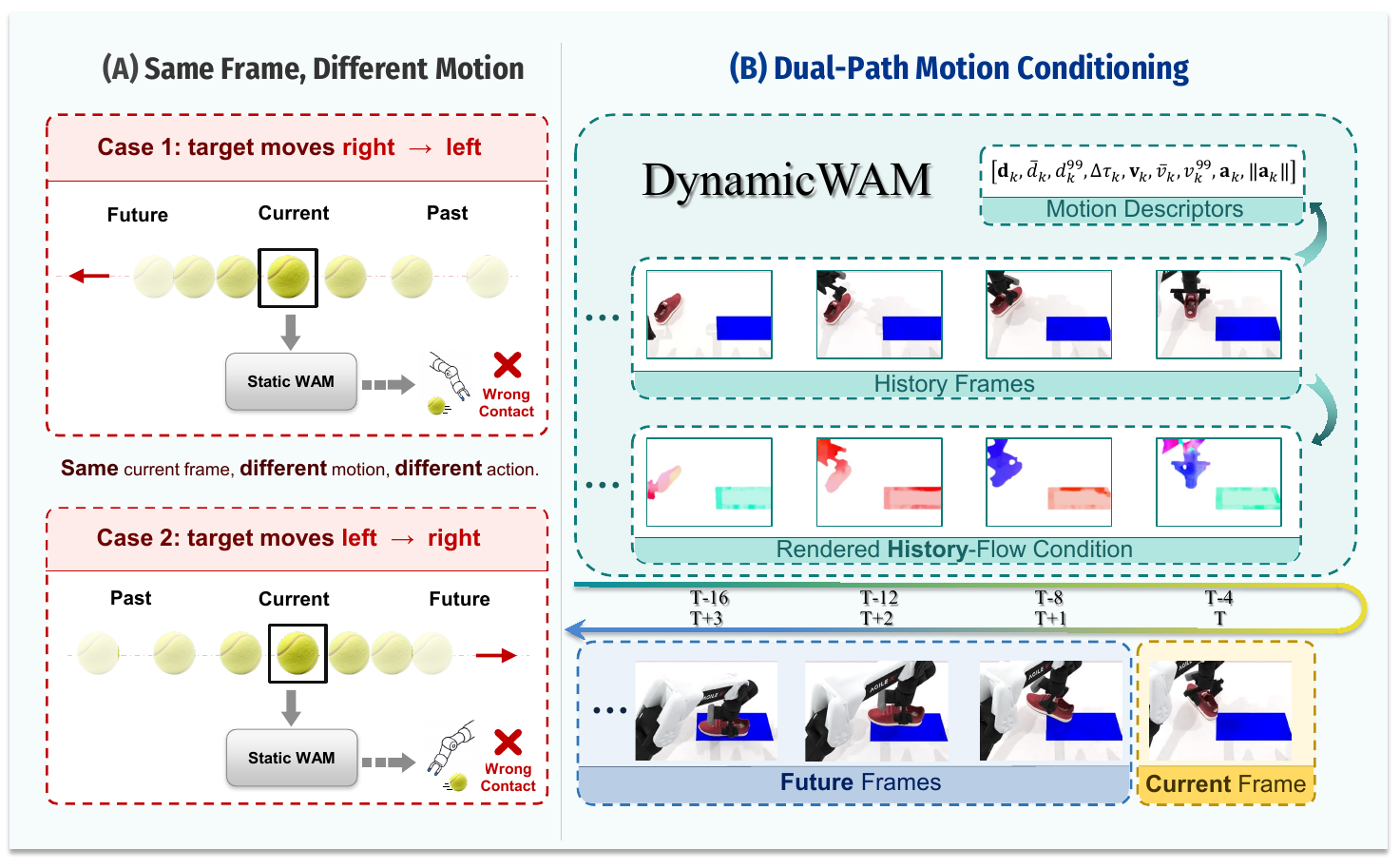}
    \caption{
Motivation. Similar current frames can correspond to opposite target motions, while
dual-path motion conditioning resolves this ambiguity using complementary
history-flow and kinematic cues.
}
    \label{fig:teaser}
\end{figure}

World--Action Models (WAMs) offer a promising framework for dynamic manipulation by coupling future visual prediction with action generation~\cite{kim2026cosmos,li2026efficient}. Their video-generation priors provide useful geometric and contact-related cues, and recent approaches reduce the computational cost of future prediction through compact video branches, low-resolution future latents, asymmetric denoising, and foresight distillation. However, most existing WAMs predict future observations primarily from the current frame and invoke their backbones synchronously at action-chunk boundaries. Consequently, as illustrated in Figure~\ref{fig:teaser}, visually similar
observations may arise from different target motions and require different
actions, while predictions valid at inference time may become outdated
before the corresponding action chunk is completed.

Recent dynamic-manipulation policies have demonstrated the benefits of temporal history, future-state prediction, and continuous action generation~\cite{fang2026towards,xie2026dynamicvla}. Nevertheless, incorporating motion into a video-based WAM raises a specific representation problem. Optical flow naturally describes where motion occurs and how it is organized in image space, making it compatible with pretrained video encoders. Yet optical-flow frames are commonly rendered with per-frame magnitude normalization, which removes their global displacement scale. Moreover, a rendered flow image does not specify the duration over which the displacement was measured. Thus, rendered flow alone cannot determine how far or how fast the target is moving, although this information is essential for determining when the robot should make contact with the moving target.

We propose DynamicWAM, a compact World--Action Model with dual-path motion conditioning for dynamic manipulation. The first path provides dense spatial motion information. For each policy query, DynamicWAM computes a temporally aligned history of optical flow from the fixed head camera, concatenates the rendered flow frames with the current observation, and encodes them using a frozen pretrained video VAE. This design allows recent motion to condition future visual prediction through the same latent pathway as visual appearance, without introducing a separate motion encoder or changing the robot action space.

The second path provides the magnitude and timing information that flow rendering discards. For each motion interval, DynamicWAM computes a kinematic descriptor containing displacement, interval duration, velocity, and acceleration. These descriptors are encoded as kinematic tokens and injected into the action expert. The rendered flow frames preserve the spatial distribution, direction, and temporal ordering of motion, whereas the kinematic tokens provide its displacement scale and temporal rate. The two paths are integrated through layer-wise joint world--action attention, allowing future prediction and action generation to use both forms of motion evidence.

To support responsive deployment, DynamicWAM uses a compact video expert distilled from a pretrained video model. We evaluate the motion-conditioning design on DOMINO under its native synchronous protocol. Adding dual-path motion conditioning improves the static WAM from 22.7\% to 38.2\% success, with manipulation score increasing from 38.3 to 53.2. For real-robot deployment, we combine DynamicWAM with Real-Time Chunking (RTC)~\cite{black2025rtc}, which predicts the next action chunk while the current chunk is being executed. Across 12 tasks covering linear, circular and compound target motion, DynamicWAM achieves a 46.7\% average success rate, exceeding the strongest evaluated baseline by 22.9 percentage points.

Our contributions are threefold:
\begin{itemize}
\item We introduce history-flow conditioning for World--Action Models, in which temporally aligned optical-flow frames are encoded together with the current observation through a frozen pretrained video VAE, providing dense recent-motion evidence for future prediction without modifying the robot action space.

\item We introduce kinematic token conditioning for World--Action Models, in which kinematic descriptors of displacement, interval duration, velocity, and acceleration are projected into tokens and injected into the action expert, providing motion magnitude and timing that complement the spatial structure encoded by the history-flow pathway.

\item We develop a compact motion-conditioning WAM and validate it on both DOMINO and a 12-task real-robot evaluation. DynamicWAM achieves 38.2\% success and a 53.2 manipulation score on DOMINO, and 46.7\% average success in real-world dynamic manipulation.
\end{itemize}

\section{Related Work}
\paragraph{Vision--Language--Action and World--Action Models.}
Vision--language--action (VLA) models learn generalist policies that map observations, language instructions, and proprioceptive states to robot actions~\cite{zitkovich2023rt,kim2024openvla,black2024pi_0, ye2026datapyramidembodiedmanipulation}. Recent work improves their scalability and action generation through large-scale robot data, generative action modeling, and efficient decoders~\cite{liu2025rdt,black2025pi05,chen2025internvlam1,hu2026spectral}. However, most VLAs predict actions directly from observed context without explicitly modeling future scene evolution~\cite{ye2025token}. World--Action Models (WAMs) address this limitation by coupling future visual prediction with action generation~\cite{kim2026cosmos,lou2026dreamtac}. Subsequent methods reduce prediction cost using compact video experts, compressed future latents, asymmetric denoising, and foresight distillation~\cite{yuan2026fast,li2026efficient,ma2026internvlaa15}. Unlike prior WAMs that focus on how predicted futures guide actions, DynamicWAM conditions both future prediction and control on recent target motion.

\paragraph{Motion Representations for Robot Control.}
Robot policies represent visual motion using dense optical flow, pixel displacement fields, or sparse point trajectories. These representations have supported motion prediction \cite{yuan2026adaptive}, visual servoing \cite{qi2026air}, flow-to-action learning \cite{yuan2026peral}, and motion-aware control~\cite{fang2025robotic,ranasinghe2026future,hu2024video,xu2024flow,wen2023any,ranasinghe2025pixel,lou2026mask,nguyen2026pixel}. Closest to our setting, PUMA incorporates historical optical flow into a VLA for short-horizon object-state prediction~\cite{fang2026towards}, while FlowWAM uses generated optical flow as an action representation or video-generation condition~\cite{chen2026flowwam}. In contrast, DynamicWAM conditions the pretrained video pathway on observed flow history and the action pathway on kinematic information.

\paragraph{Dynamic Manipulation and Responsive Execution.}
Manipulation of moving targets has been studied in robotic soccer, table tennis, human--robot handover, and reactive grasping~\cite{kitano1997robocup,dambrosio2025achieving,christen2023learning,liu2023target,zhang2025dynamic}. Many such systems rely on task-specific perception, known motion structure, or specialized controllers. Recent generalist approaches broaden this setting. DOMINO provides a large-scale benchmark for manipulation under diverse target motion, and PUMA combines historical motion cues with short-horizon prediction~\cite{fang2026towards}. DynamicVLA improves responsiveness through a compact architecture, continuous inference, and latency-aware action streaming~\cite{xie2026dynamicvla}. Complementary execution methods such as Real-Time Chunking generate the next action chunk while the current one is being executed, and adjust the guidance weight to ensure action chunks are renewed smoothly, avoiding pauses caused by model inference~\cite{black2025rtc}. 
DynamicWAM enables motion-conditioning future prediction in a WAM and responsive real-robot deployment through asynchronous chunk execution.

\begin{figure*}[t]
    \centering
    \includegraphics[width=0.95\textwidth]{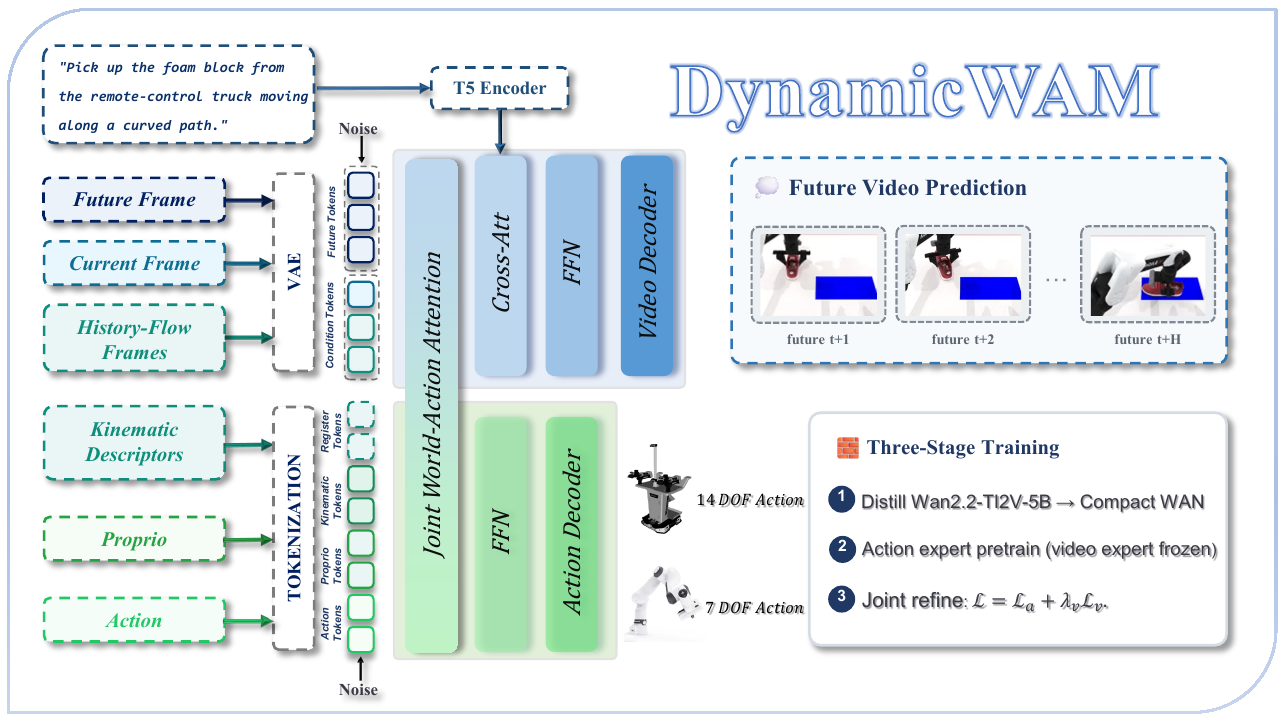}
    \caption{
    Overview of Architecture. Temporally aligned history-flow frames are
    encoded with the current multi-view observation through a frozen
    pretrained video VAE to provide spatial motion cues, while kinematic
    descriptors of displacement, duration, velocity, and acceleration are
    injected into the action expert to provide metric scale and timing.
    The video and action experts interact through joint world--action
    attention for coupled future-video prediction and action generation.
    Training proceeds through video-expert distillation, action-expert
    pretraining, and joint refinement.
    }
    \label{fig:dynamicwam-overview}
\end{figure*}

\section{Method}

DynamicWAM extends a standard World--Action Model with dual-path motion conditioning. As shown in Figure~\ref{fig:dynamicwam-overview}, history-flow frames are encoded with the current observation through a frozen video VAE, while kinematic descriptors are injected into the action expert. The two paths are coupled through layer-wise joint world--action attention.

\subsection{World--Action Backbone}

We consider language-conditioned manipulation in scenes that continue to evolve while the robot acts. At each action-chunk boundary \(t\), the policy receives the current multi-view observation \(O_t\), a language instruction \(\ell\), and the proprioceptive state \(s_t\in\mathbb{R}^{14}\) or \(\mathbb{R}^{7}\), and predicts an action chunk
\begin{equation}
A_t=a_{t:t+H_a-1},
\end{equation}
where \(H_a=16\).

The backbone comprises a compact video expert distilled from Wan2.2-TI2V-5B~\cite{wan2025wan} and an action expert that predicts joint-position actions. The pretrained video VAE is kept frozen. Within each transformer block, video and action queries attend to the concatenated keys and values of both streams (joint world--action attention).

Let \(Z_t^{\sigma_v}\) and \(A_t^{\sigma_a}\) denote noisy future visual latents and actions at noise levels \(\sigma_v\) and \(\sigma_a\). The coupled model is
\begin{equation}
\label{eq:field}
(\hat{u}^{v},\hat{u}^{a})
=
F_{\Theta}
\left(
Z_t^{\sigma_v},
A_t^{\sigma_a};
\mathcal{C}
\right),
\qquad
\mathcal{C}
=
(O_t,\ell,s_t).
\end{equation}
Here, \(F_{\Theta}\) is the coupled world--action vector-field predictor with parameters \(\Theta\), and \(\hat{u}^{v}\) and \(\hat{u}^{a}\) denote its predicted conditional-flow-matching vector fields for the visual-latent and action streams, respectively. Under \(\mathcal{C}\) alone, visually similar observations can arise from distinct target motions and thus demand different contact actions.

\subsection{Dual-Path Motion Conditioning}

\begin{figure}[t]
    \centering
    \includegraphics[width=\columnwidth]{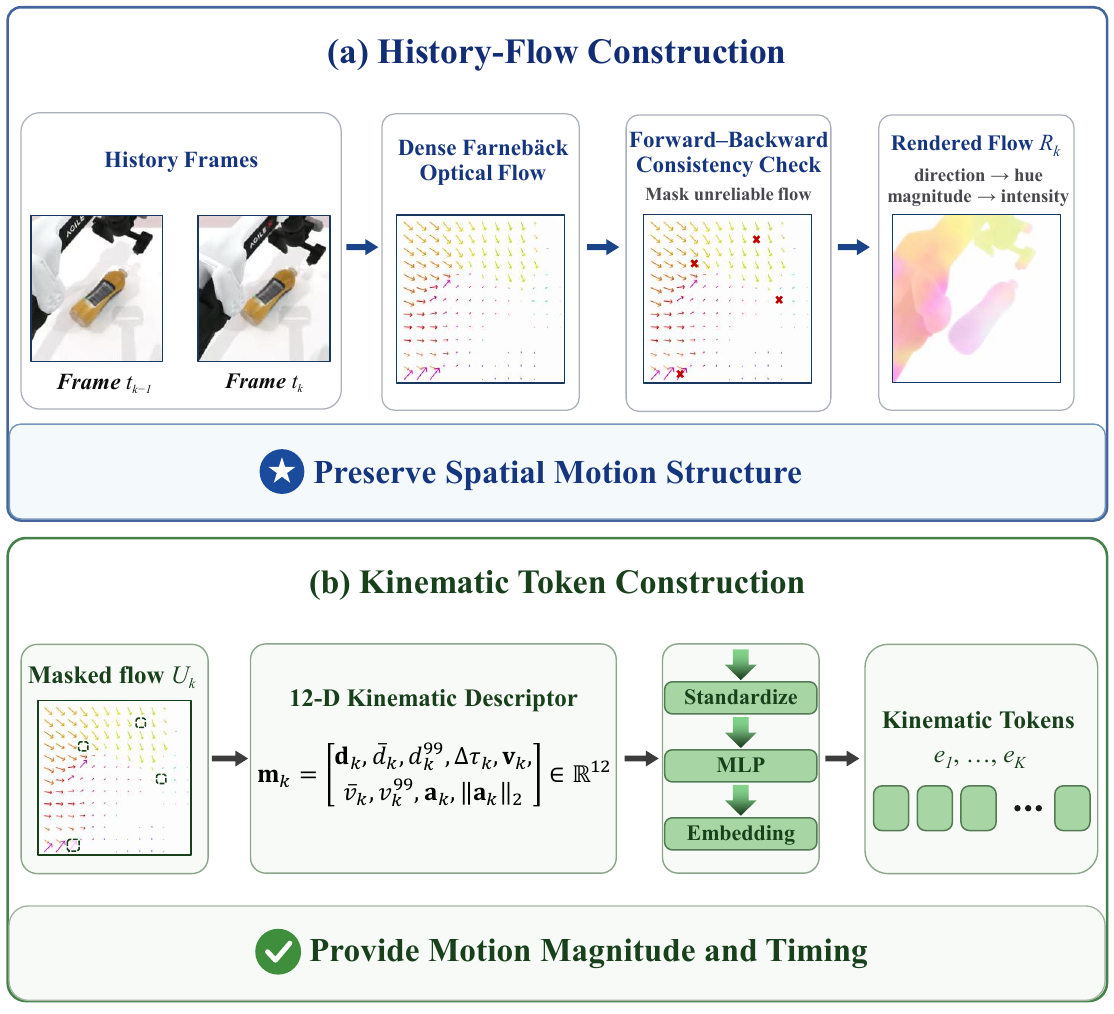}
    \caption{
    Dual-path motion conditioning. (a) Each history interval is converted into a masked optical-flow field \(\hat{U}_k\) and rendered as an RGB image \(R_k\), providing spatial motion structure to the video pathway. (b) Statistics of the same masked flow field, together with the interval duration \(\Delta\tau_k\), form a 12-D descriptor \(\mathbf{m}_k\) that is mapped to a kinematic token \(e_k\), providing motion magnitude and timing to the action pathway.
    }
    \label{fig:dual-path-motion}
\end{figure}

DynamicWAM conditions the WAM on recent target motion through two complementary paths, as illustrated in Figure~\ref{fig:dual-path-motion}. History-flow frames preserve the spatial organization of motion and are encoded with the current observation by the frozen video VAE. Kinematic tokens provide the displacement scale and timing information absent from per-frame flow renderings and are injected into the action expert. Starting from the base condition \(\mathcal{C}=(O_t,\ell,s_t)\) in Eq.~\eqref{eq:field}, DynamicWAM augments the coupled model with two additional conditioning sources: the rendered flow history \(\mathcal{R}_t\) through the video pathway and kinematic tokens derived from \(\mathcal{M}_t\) through the action pathway.

\subsubsection{History-Flow Conditioning}

To encode recent scene motion, we partition the observation history at chunk boundary \(t\) into \(K=4\) intervals with endpoints separated by \(\Delta=4\) policy steps:
\begin{equation}
\label{eq:history-endpoints}
n_k
=
t-(K-k)\Delta,
\qquad
k=0,\ldots,K.
\end{equation}
Each endpoint carries a timestamp \(\tau_k\); intervals missing at episode start are marked invalid.

For each interval, we compute dense Farneb\"ack optical flow~\cite{farneback2003two} from the fixed head camera and apply a forward--backward consistency check, yielding the masked flow field \(\hat{U}_k\). Each field is rendered as an RGB image \(R_k=\rho(\hat{U}_k)\), with direction mapped to hue and magnitude normalized by the frame-specific 99th-percentile \(q(\hat{U}_k)\):
\begin{equation}
\label{eq:flow-rendering}
r_k(p)
=
\min\!\left(
\frac{\lVert\hat{U}_k(p)\rVert_2}{q(\hat{U}_k)},\,
1
\right).
\end{equation}

The rendered history \(\mathcal{R}_t=(R_1,\ldots,R_K)\) is concatenated with the current observation along the temporal axis and encoded by the frozen pretrained video VAE:
\begin{equation}
\label{eq:vae}
Z_t^{c}
=
E_{\mathrm{VAE}}
\left(
[R_1,\ldots,R_K,O_t]
\right),
\end{equation}
yielding
\begin{equation}
\label{eq:flow-condition}
\mathcal{C}_{\mathrm{flow}}
=
(O_t,\ell,s_t,\mathcal{R}_t).
\end{equation}
Unlike methods that treat flow as an action representation~\cite{chen2026flowwam}, we condition on observed flow while leaving the robot action space unchanged.

\subsubsection{Scale and Timing Ambiguity of Flow Rendering}

Per-frame normalization removes absolute displacement scale: for \(\alpha>0\),
\begin{equation}
\label{eq:invariance}
\rho(\alpha U)=\rho(U)
\end{equation}
up to masking and quantization. Fields with the same direction and spatial pattern but different magnitudes can therefore yield nearly identical renderings. The rendered image also omits the elapsed time \(\Delta\tau_k\) between endpoint frames, so velocity cannot be recovered from \(\mathcal{R}_t\) alone. The kinematic path below supplies these missing magnitude and timing cues without replacing the flow representation.

\subsubsection{Kinematic Token Conditioning}

For interval \(k\), let \(\Omega_k\) denote the set of pixels that remain valid after the forward--backward consistency check. If \(\Omega_k=\varnothing\), the interval is marked invalid; otherwise, we compute the mean signed displacement and mean displacement magnitude over these valid pixels as
\begin{equation}
\label{eq:stats}
\mathbf{d}_k
=
\frac{1}{|\Omega_k|}
\sum_{p\in\Omega_k}
\hat{U}_k(p),
\qquad
\bar{d}_k
=
\frac{1}{|\Omega_k|}
\sum_{p\in\Omega_k}
\left\lVert
\hat{U}_k(p)
\right\rVert_2,
\end{equation}
together with the 99th-percentile magnitude \(d_k^{99}=q(\hat{U}_k)\), computed over \(\Omega_k\), and the interval duration \(\Delta\tau_k=\tau_k-\tau_{k-1}\). Velocity statistics follow by normalization with \(\Delta\tau_k\):
\begin{equation}
\label{eq:velocity}
\mathbf{v}_k
=
\frac{\mathbf{d}_k}{\Delta\tau_k},
\qquad
\bar{v}_k
=
\frac{\bar{d}_k}{\Delta\tau_k},
\qquad
v_k^{99}
=
\frac{d_k^{99}}{\Delta\tau_k}.
\end{equation}
With interval centers \(c_k=(\tau_{k-1}+\tau_k)/2\), acceleration is estimated as
\begin{equation}
\label{eq:acceleration}
\mathbf{a}_k
=
\frac{\mathbf{v}_k-\mathbf{v}_{k-1}}{c_k-c_{k-1}}.
\end{equation}
The resulting descriptor is
\begin{equation}
\label{eq:descriptor}
\mathbf{m}_k
=
\bigl[
\mathbf{d}_k,\,
\bar{d}_k,\,
d_k^{99},\,
\Delta\tau_k,\,
\mathbf{v}_k,\,
\bar{v}_k,\,
v_k^{99},\,
\mathbf{a}_k,\,
\lVert\mathbf{a}_k\rVert_2
\bigr]
\in\mathbb{R}^{12}.
\end{equation}

Descriptors are standardized with dataset-level statistics. Invalid intervals use a learned invalid-interval embedding; when a predecessor is unavailable, acceleration dimensions are masked and replaced by a learned invalid-acceleration embedding. Each \(\mathbf{m}_k\) is projected to the action-expert width by a two-layer MLP and combined with interval-position and type embeddings, producing kinematic tokens \(e_1,\ldots,e_K\). They are inserted into the action stream as
\begin{equation}
\label{eq:layout}
X^a
=
\bigl[
e_s(s_t);\,
e_a(A_t^{\sigma_a});\,
e_1,\ldots,e_K;\,
g_1,\ldots,g_4
\bigr].
\end{equation}
The tokens participate in every joint-attention block; the action decoder reads only the original state--action tokens, so the robot action representation remains unchanged. The full condition is
\begin{equation}
\label{eq:full-condition}
\mathcal{C}_{\mathrm{full}}
=
(O_t,\ell,s_t,\mathcal{R}_t,\mathcal{M}_t),
\end{equation}
where \(\mathcal{M}_t\) collects the kinematic descriptors and validity indicators.

\subsection{Joint Training Objective}

During joint refinement, both streams are optimized with conditional flow
matching. For each \(q\in\{v,a\}\), given a clean target \(x_q\), noise
\(\epsilon_q\sim\mathcal{N}(0,I)\), and noise level
\(\sigma_q\in[0,1]\), we define
\begin{equation}
x_q^{\sigma_q}
=
(1-\sigma_q)x_q+\sigma_q\epsilon_q,
\qquad
u_q^\star=\epsilon_q-x_q.
\end{equation}
The joint objective is
\begin{equation}
\mathcal{L}
=
\mathbb{E}\!\left[
\|\hat{u}^{a}-u_a^\star\|_2^2
\right]
+
\lambda_v
\mathbb{E}\!\left[
\|\hat{u}^{v}-u_v^\star\|_2^2
\right].
\end{equation}
Details of distillation, pretraining, joint refinement, and inference are
provided in the appendix.

\subsection{RTC-Based Asynchronous Execution}

For responsive real-robot deployment, DynamicWAM adopts Real-Time
Chunking (RTC)~\cite{black2025rtc}. Instead of waiting for the current
action chunk to finish before issuing the next policy query, RTC predicts
the next chunk while the current one is being executed and smoothly
merges consecutive chunks. RTC does not reduce the latency of an
individual model query; rather, it hides inference latency through
execution--inference overlap, reducing pauses and perception--execution
lag. RTC is used only for real-robot deployment, while DOMINO evaluation
retains the benchmark's native synchronous protocol.

\section{Simulation Experiments}
\label{sec:experiments}

Our simulation experiments address two questions. \textbf{Q1:} Does history-flow conditioning improve dynamic manipulation by providing the spatial structure of recent motion? \textbf{Q2:} Do kinematic tokens provide complementary gains by supplying motion magnitude and timing?

\subsection{Experimental Setup}
\label{sec:simulation-setup}

We evaluate DynamicWAM on all 35 tasks of DOMINO Level~1 under the clean dynamic setting, using 100 accepted episodes per task with unseen instructions and a fixed evaluation seed. All variants follow the benchmark's native synchronous protocol, predicting 16 joint-position actions each time without background inference or action merging. Models are trained on 10{,}500 demonstrations, comprising 150 clean and 150 randomized episodes per task. We report success rate (SR) and manipulation score (MS).

\subsection{Main Results}
\label{sec:main-results}

\begin{table}[!htbp]
\centering
\small
\setlength{\tabcolsep}{3pt}
\begin{tabular}{lccc}
\toprule
Method
& \begin{tabular}[b]{@{}c@{}}Latency\\(ms)$\downarrow$\end{tabular}
& \begin{tabular}[b]{@{}c@{}}SR\\(\%)$\uparrow$\end{tabular}
& MS$\uparrow$ \\
\midrule
\multicolumn{4}{l}{\emph{Fine-tuned baselines}} \\
OpenVLA~\cite{kim2024openvla} & 173.6 & 1.5 & 6.1 \\
$\pi_0$-FAST~\cite{pertsch2025fast} & 119.9 & 3.5 & 20.9 \\
VLA-Adapter~\cite{wang2025vlaadapter} & 83.6 & 4.4 & 24.3 \\
RDT-1B~\cite{liu2025rdt} & 246.3 & 5.3 & 17.7 \\
InternVLA-M1~\cite{chen2025internvlam1} & 183.8 & 5.4 & 27.6 \\
Isaac-GR00T~\cite{bjorck2025gr00t} & 76.4 & 6.1 & 28.6 \\
$\pi_0$~\cite{black2024pi_0} & 52.9 & 8.2 & 24.0 \\
OpenVLA-OFT~\cite{kim2025oft} & 83.1 & 9.1 & 24.1 \\
$\pi_{0.5}$~\cite{black2025pi05} & 59.1 & 9.6 & 26.2 \\
PUMA~\cite{fang2026towards} & 85.4 & 17.2 & 35.0 \\
InternVLA-A1.5~\cite{ma2026internvlaa15}
& 158.4 & \underline{29.3} & \underline{42.5} \\
DynamicWAM (ours) & 173.7 & \textbf{38.2} & \textbf{53.2} \\
\bottomrule
\end{tabular}
\caption{
Performance on DOMINO Level~1 under the clean dynamic setting. Latency is measured per policy query; SR and MS denote success rate and
manipulation score.
}
\label{tab:main}
\end{table}

Table~\ref{tab:main} compares DynamicWAM with existing methods on
DOMINO. DynamicWAM achieves 38.2\% SR and 53.2 MS, outperforming all
evaluated baselines on both metrics. It improves over the strongest
baseline, InternVLA-A1.5, by 8.9 SR points and 10.7 MS points, and
outperforms PUMA by 21.0 SR points and 18.2 MS points.

DynamicWAM achieves the best overall performance with a latency of
173.7\,ms per policy query. Although it is not the fastest method, its
inference latency remains competitive and within the same order of
magnitude as the fastest baselines. Meanwhile, it is approximately
3.2$\times$ faster than InternVLA-A1.5, the strongest competing method,
while improving SR and MS by 8.9 and 10.7 points, respectively. This
result demonstrates a favorable trade-off between inference efficiency
and dynamic-manipulation performance.

\subsection{Ablation Study}
\label{sec:ablations}

\begin{figure}[t]
\centering
\includegraphics[width=\columnwidth]
{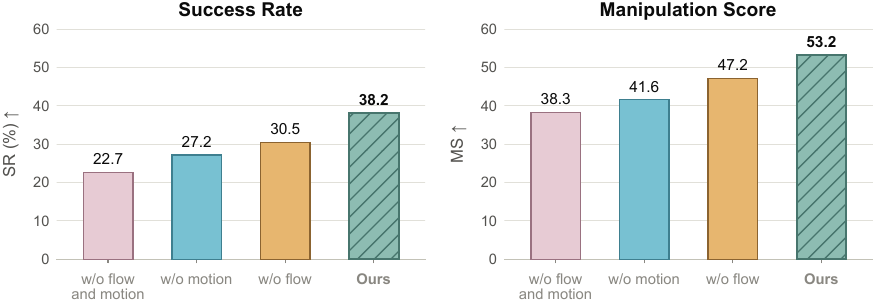}
\caption{
Ablation of dual-path motion conditioning on DOMINO Level~1. History flow and kinematic tokens provide complementary gains in both
SR and MS.
}
\label{fig:ablation}
\end{figure}
\begin{figure}[t]
\centering
\includegraphics[width=\columnwidth]{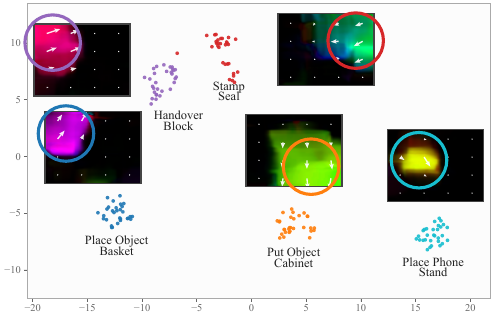}
\caption{
t-SNE visualization of history-flow latents. Samples from five representative tasks exhibit locally coherent motion
structures in the frozen video-VAE latent space.
}
\label{fig:flow-tsne}
\end{figure}

Figure~\ref{fig:ablation} isolates the contributions of the two
motion-conditioning paths. Adding history-flow conditioning to
DynamicWAM (w/o flow and motion) improves SR from 22.7\% to 27.2\%
and MS from 38.3 to 41.6. This result supports Q1: preserving the
spatial structure and temporal ordering of recent motion benefits
dynamic manipulation.

The kinematic-token-only variant, DynamicWAM (w/o flow), achieves
30.5\% SR and 47.2 MS, indicating that motion magnitude and timing
provide information not retained by normalized flow rendering.
Combining both paths yields 38.2\% SR and 53.2 MS, outperforming the
stronger single-path variant by 7.7 SR points and 6.0 MS points.
These results support Q2 and demonstrate that the two paths are
complementary rather than redundant.

% \begin{figure}[t]
% \centering
% \includegraphics[width=\columnwidth]{Figures/t-sne_column.pdf}
% \caption{t-SNE visualization of flow latents obtained by encoding rendered history-flow images with the frozen pretrained video VAE (Eq.~\ref{eq:vae}), shown for five representative DOMINO tasks with 30 episodes per task and one rendered flow frame per episode. A representative rendered flow image is shown next to each cluster, with the ring color matching the cluster and marking the moving region. Episodes from the same task form compact clusters whose flow patterns match the task's characteristic target motion.}
% \label{fig:flow-tsne}
% \end{figure}

\subsection{Analysis}
\label{sec:further-analysis}

\subsubsection{Task-Level Effects}

The largest gains occur on tasks requiring accurately timed contact.
On \mbox{move\_pillbottle\_pad}, success increases from 42\% for
DynamicWAM (w/o flow and motion) to 55\% with history flow and 91\%
with the full model. On \mbox{beat\_block\_hammer}, it increases from
36\% to 49\% and 84\%, respectively. Kinematic tokens are particularly
important on \mbox{place\_phone\_stand}, where history flow alone
achieves 32\% success, compared with 80\% for the full model. The gains
are not uniform: the full model trails the history-flow-only variant by
10 points on \mbox{put\_object\_cabinet} and by 8 points on
\mbox{place\_can\_basket}. Complete results are provided in the appendix.

\subsubsection{Motion Structure in Flow Latents}

Figure~\ref{fig:flow-tsne} visualizes frozen-VAE latents of rendered
history-flow frames from five representative tasks. The samples form
locally coherent groups associated with differences in motion direction,
spatial extent, and image location, suggesting that the pretrained VAE
preserves motion-relevant spatial variation. The distributions remain
multimodal and partially overlapping across all tasks; thus, the
visualization provides qualitative evidence of local motion structure
rather than global task separability.

\begin{table}[!htbp]
\centering
\begingroup
\small
\setlength{\tabcolsep}{1mm}
\begin{tabular}{lrrrr}
\toprule
Method & L1 & L2 & L3 & Avg. \\
\midrule
\multicolumn{5}{l}{\emph{External baselines}} \\
InternVLA-A1.5 & 0.00 & 0.00 & 0.00 & 0.00 \\
$\pi_{0.5}$ & 32.50 & 38.75 & 0.00 & 23.75 \\
DynamicVLA & 30.00 & 33.75 & 0.00 & 21.25 \\
\midrule
\multicolumn{5}{l}{\emph{DynamicWAM variants}} \\
DynamicWAM (w/o flow \& motion)
& 37.50 & 35.00 & 2.50 & 25.00 \\
DynamicWAM (w/o motion)
& 51.25 & 42.50 & 15.00 & 36.25 \\
DynamicWAM (w/o flow)
& 55.00 & 47.50 & 17.50 & 40.00 \\
DynamicWAM (ours)
& \textbf{70.00} & \textbf{51.25}
& \textbf{18.75} & \textbf{46.67} \\
\bottomrule
\end{tabular}
\endgroup
\caption{
Real-world success rates (\%). L1--L3 denote linear, circular, and compound target motion.
}
\label{tab:real-world-performance}
\end{table}

\section{Real-World Experiments}
\label{sec:real-world}

We evaluate DynamicWAM in the real world from three perspectives:
overall performance across different target-motion geometries, the
contributions of dual-path motion conditioning and RTC-based
asynchronous execution, and robustness to visual distribution shifts.

\subsection{Tasks and Experimental Setup}
\label{sec:real-world-setting}
% \begin{figure*}[t]
%     \centering
%     \includegraphics[width=1\textwidth, trim=105 150 120 10, clip]{Figures/realworld.pdf}
%     \caption{\textbf{Real-World Simulation.} Representative real-world dynamic manipulation tasks across three levels of motion complexity. From top to bottom, the robot grasps a tennis ball undergoing linear motion along an inclined rail (Level 1), a snack undergoing uniform circular motion on a rotating turntable (Level 2), and a cube carried along a curved trajectory by a moving vehicle (Level 3). Columns show the temporal progression from the initial observation at \(T\) to the completed grasp at \(T+N\).
%     }
%     \label{fig:real-world-tasks}
% \end{figure*}

We construct a real-robot evaluation suite containing 12 tasks grouped
into three levels according to target-motion geometry, with four tasks
per level. Level~1 covers linear motion generated by a remote-controlled
vehicle, an inclined rail, a conveyor belt, or a self-propelled toy.
Level~2 considers uniform circular motion on a rotating turntable.
Level~3 includes compound motion, such as random bouncing,
simultaneous translation and rotation, and curved trajectories.

\begin{figure}[t]
\centering
\includegraphics[
    width=\columnwidth
]{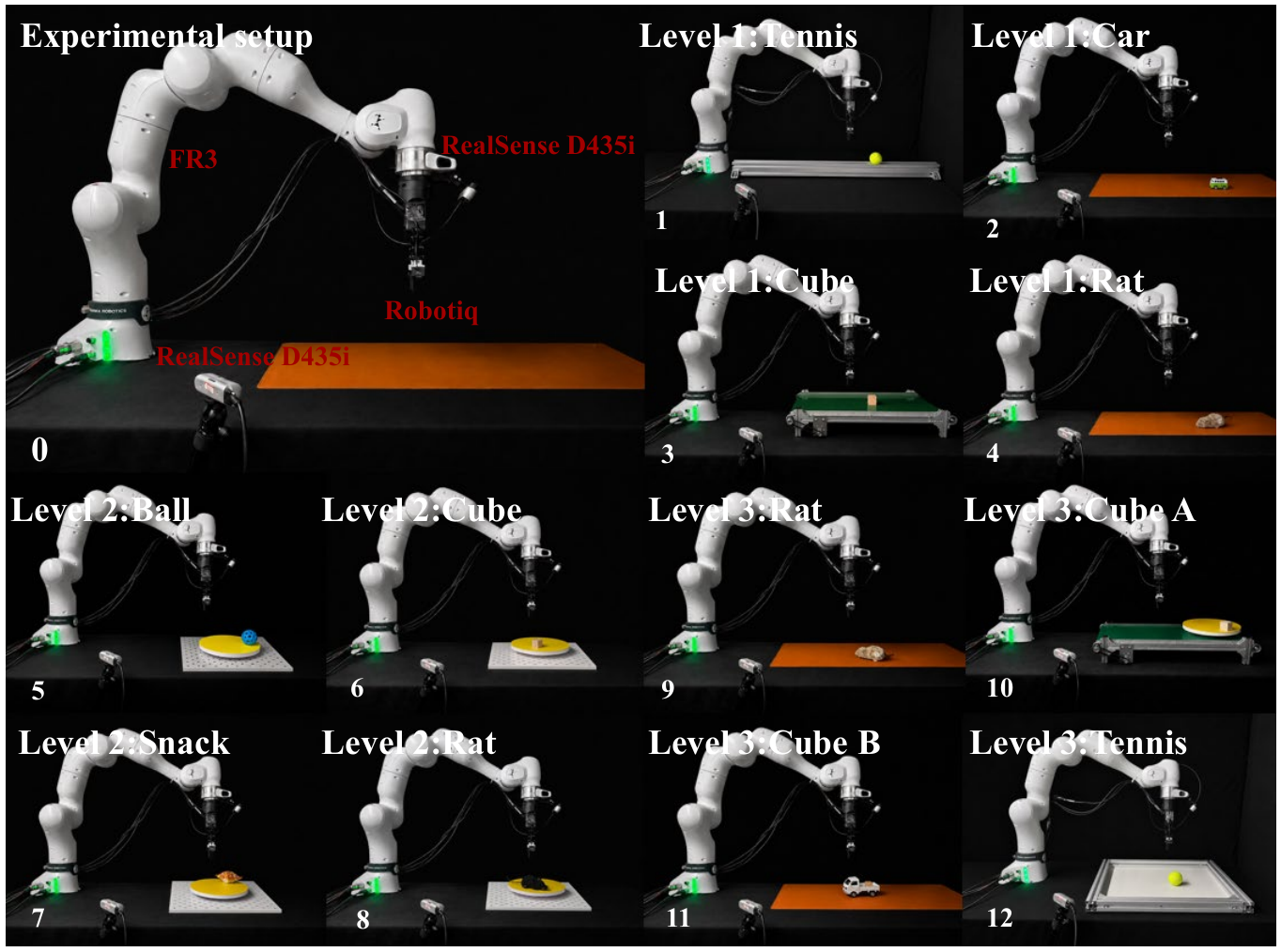}
\caption{
Real-world evaluation setup. Twelve tasks span three levels of target-motion complexity and two
Franka Research 3 gripper configurations.
}
\label{fig:real-world-setting}
\end{figure}

\begin{figure*}[t]
\centering
\includegraphics[
    width=0.95\textwidth
]{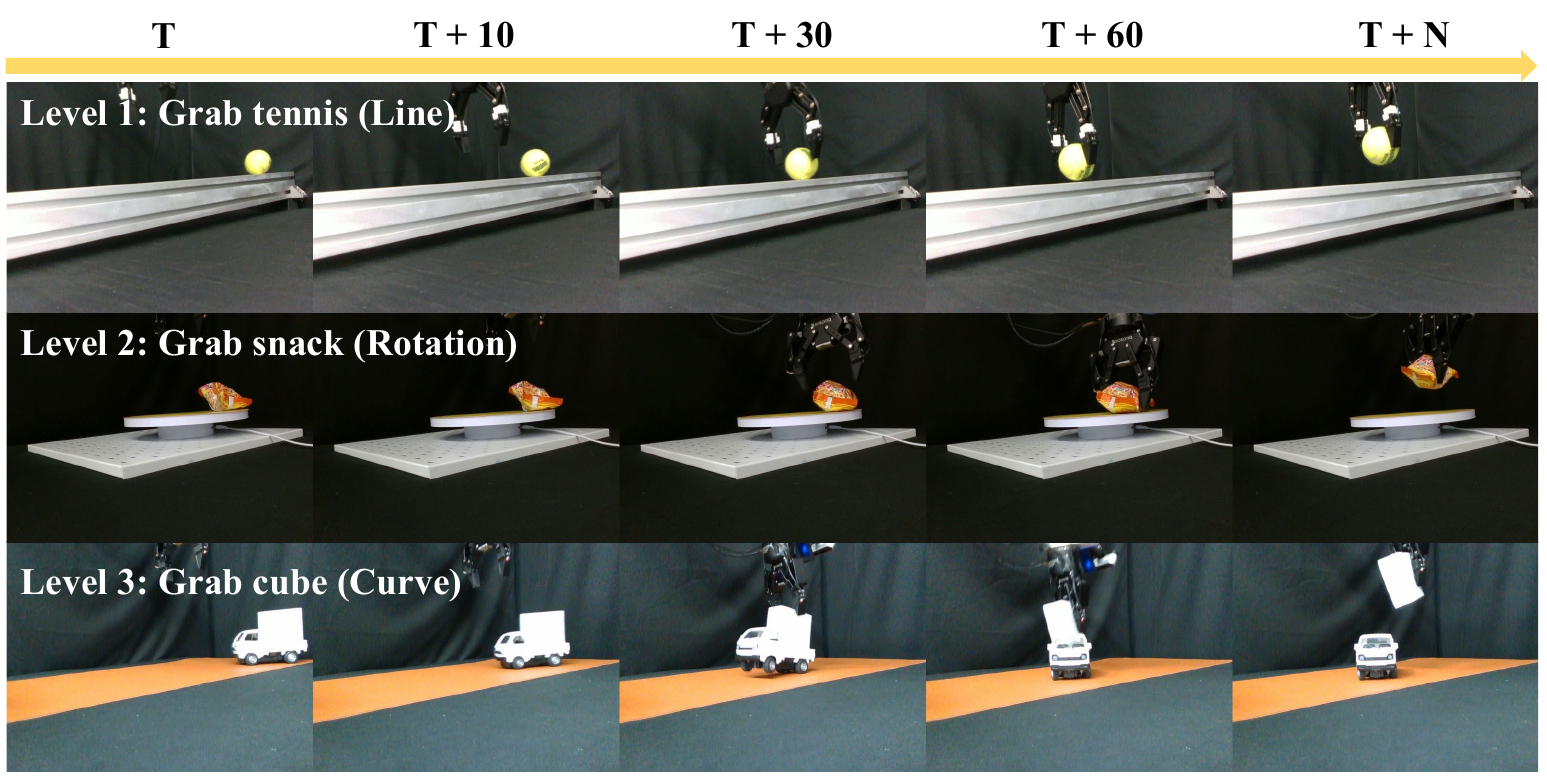}
\caption{
Representative real-world tasks. From top to bottom, the targets undergo linear, circular, and compound motion (Levels~1--3). Columns show execution from the initial
observation at \(T\) to the completed grasp at \(T+N\).
}
\label{fig:real-world-tasks}
\end{figure*}

All policies are deployed on Franka Research 3 manipulators equipped
with either a CTAG2F90 or a Robotiq 2F-140 parallel-jaw gripper, and visual observations are provided by two Intel RealSense D435i cameras.
Figure~\ref{fig:real-world-setting} and Figure~\ref{fig:real-world-tasks}
show the hardware configurations and representative task executions.

\subsection{Experimental Protocol}
\label{sec:real-world-protocol}

Each method is initialized from its corresponding pretrained checkpoint
and fine-tuned on the same 1{,}200-demonstration dataset, containing
100 demonstrations per task. We evaluate each policy for 20 trials/task, yielding 240 trials/method.

We compare DynamicWAM with InternVLA-A1.5~\cite{ma2026internvlaa15},
\(\pi_{0.5}\)~\cite{black2025pi05}, and
DynamicVLA~\cite{xie2026dynamicvla}. The external baselines follow
their official synchronous deployment protocols. DynamicWAM uses
Real-Time Chunking (RTC)~\cite{black2025rtc} in the main comparison,
generating the next action chunk while the current chunk is being
executed. We report success rates for each motion level and their average
across all 12 tasks.

\subsection{Real-World Results}
\label{sec:real-world-results}

Table~\ref{tab:real-world-performance} shows that DynamicWAM achieves
the highest success rate at all three motion levels. Its average success
rate is 46.67\%, exceeding the strongest external baseline,
\(\pi_{0.5}\), by 22.92 percentage points. The improvement is largest
on Level~1, where DynamicWAM reaches 70.00\% success compared with
32.50\% for \(\pi_{0.5}\) and 30.00\% for DynamicVLA. DynamicWAM also
achieves 51.25\% on Level~2 and remains effective on the substantially
more challenging Level~3 tasks, where the external baselines record no
successful trials.

The DynamicWAM variants exhibit the same ordering as in simulation.
History-flow conditioning and kinematic-token conditioning independently
increase the average success rate from 25.00\% to 36.25\% and 40.00\%,
respectively. Combining both paths further improves performance to
46.67\%, confirming that spatial motion structure and motion magnitude
and timing remain complementary in real-world deployment.

\subsection{Effect of Asynchronous Execution}
\label{sec:rtc-ablation}

Table~\ref{tab:rtc-ablation} isolates the effect of asynchronous
execution. RTC improves the average success rate from 42.08\% to
46.67\% (+4.59 points), with gains of 2.50, 5.00, and 6.25 points on
Levels~1--3, respectively. The larger gain on Level~3 suggests that RTC
is particularly valuable under complex target motion. The synchronous
variant nevertheless remains stronger than all external baselines,
confirming that motion conditioning and asynchronous execution provide
complementary benefits.

\subsection{Visual Generalization}
\label{sec:visual-generalization}

\begin{table}[!htbp]
\centering
\begingroup
\small
\setlength{\tabcolsep}{1mm}
\begin{tabular}{lrrrr}
\toprule
Method & L1 & L2 & L3 & Avg. \\
\midrule
DynamicWAM (w/o RTC)
& 67.50 & 46.25 & 12.5 & 42.08 \\
DynamicWAM (w/ RTC)
& \textbf{70.00} & \textbf{51.25}
& \textbf{18.75} & \textbf{46.67} \\
\bottomrule
\end{tabular}
\endgroup
\caption{
Effect of RTC on real-world success rates (\%). Both variants use the full DynamicWAM model.
}
\label{tab:rtc-ablation}
\end{table}

We further evaluate the four Level~1 tasks under background, lighting,
and object-color shifts, using 20 trials per task and condition.
As shown in Table~\ref{tab:visual-generalization}, DynamicWAM achieves
the highest success rate under every visual shift and improves OOD-SR
from 17.5\% for \(\pi_{0.5}\) and 20.0\% for the current-frame variant
to 36.3\%.

The retention ratios are similar across methods, indicating that
DynamicWAM's absolute out-of-distribution advantage primarily follows
from its stronger in-distribution manipulation capability rather than
a disproportionate increase in relative robustness. Background changes
remain the most challenging shift for all evaluated methods.

\begin{table}[!htbp]
\centering
\begingroup
\small
\setlength{\tabcolsep}{1.8pt}
\begin{tabular}{lcccccc}
\toprule
Method
& SR$_{\mathrm{ID}}$
& BG
& Light
& Color
& OOD-SR
& Retain \\
\midrule
$\pi_{0.5}$
& 32.5 & 12.5 & 17.5 & 22.5 & 17.5 & 0.54 \\
\begin{tabular}[c]{@{}l@{}}
DynamicWAM\\
(w/o flow \& motion)
\end{tabular}
& 37.5 & 6.3 & 30.0 & 23.8 & 20.0 & 0.53 \\
DynamicWAM (ours)
& \textbf{70.0}
& \textbf{25.0}
& \textbf{43.8}
& \textbf{40.0}
& \textbf{36.3}
& 0.52 \\
\bottomrule
\end{tabular}
\endgroup
\caption{
Visual generalization on Level~1 tasks. OOD-SR averages background, lighting, and object-color shifts;
Retain \(=\) OOD-SR/SR$_{\mathrm{ID}}$.
}
\label{tab:visual-generalization}
\end{table}

\section{Conclusion}

We presented DynamicWAM, a motion-conditioned World--Action Model that combines history flow and kinematic tokens to condition future prediction and action generation for dynamic manipulation. DynamicWAM achieves a 38.2\% success rate and a 53.2 manipulation score on DOMINO, as well as a 46.7\% average success rate across 12 real-world tasks with Real-Time Chunking, while ablations confirm the complementary benefits of the two motion-conditioning paths and asynchronous execution. However, the current framework relies on two-dimensional image-space motion from a fixed external camera and does not explicitly isolate the instructed target from other moving regions, making it sensitive to distractors, occlusion, and viewpoint changes. Future work will explore target-aware motion extraction and three-dimensional motion modeling to improve robustness in cluttered and more complex dynamic environments.

\bibliography{aaai2027}

\clearpage

\appendix

\section{Experimental Details}
\label{app:experimental-details}

\subsection{DOMINO Benchmark and Task Definitions}
\label{app:domino}

DOMINO is a large-scale benchmark for dynamic manipulation built on RoboTwin~2.0. It comprises 35 Level-1 tabletop tasks in which a target object undergoes scripted motion during manipulation. Each task provides expert demonstrations under both clean and randomized scene layouts. Dynamic motion is generated by a trajectory planner with three complexity levels: Level~1 uses piecewise-constant-velocity motion; Level~2 uses polynomial trajectories with time-varying speed; Level~3 composes 2--3 segments with abrupt transitions. All simulation experiments in the main paper use the clean dynamic setting with dynamic level~1 and dynamic coefficient~0.1.

We evaluate all 35 Level-1 tasks, each task is associated with a language instruction template. During evaluation we use the benchmark's unseen instruction split. Success is determined by the native DOMINO task success predicate. The manipulation score (MS) averages partial progress signals defined by the benchmark and therefore captures near-success behavior not reflected in the binary success bit.

\subsection{Real-World Task Suite}
\label{app:realworld-tasks}

We evaluate DynamicWAM on a 12-task real-robot suite spanning
three levels of target-motion complexity, with four tasks per
level and 20 trials per task, totaling 240 trials per method.
Each task requires the robot to intercept and grasp a moving
target at an appropriate point along its trajectory.

Level~1 evaluates linear target motion. In Line Tennis, a tennis
ball rolls along an inclined rail. In Line Car, a target carried
by a remote-controlled vehicle moves along a straight path with
varying headings and interception positions. In Line Cube, a
wooden cube is transported laterally by a conveyor belt. In Line
Rat, a self-propelled toy rat moves along approximately straight
paths with varying headings and grasp positions.

Level~2 evaluates uniform circular motion using four turntable
tasks: TT-Ball, TT-Cube, TT-Snack, and TT-Rat. A blue ball,
wooden cube, snack package, or toy rat is placed on a rotating
turntable. The target position continuously changes with the
turntable phase, requiring the robot to identify an appropriate
interception time and grasp location.

Level~3 evaluates curved and compound target motion. In Complex
Rat, a self-propelled toy rat follows curved trajectories with
varying motion paths and interception positions. In Complex
Cube~A, a wooden cube is carried by a rotating turntable mounted
on a moving conveyor belt, combining translational and rotational
motion. In Complex Cube~B, a cube-shaped foam block is carried
by a remote-controlled vehicle following a curved path, and the
robot must grasp the block from the moving vehicle. In Complex
Tennis, a tennis ball follows a curved trajectory inside a metal
tray.

\textbf{Hardware.}
Policies are deployed on Franka Research~3 arms with either a CTAG2F90 or a Robotiq~2F-140 parallel-jaw gripper. Observations come from two Intel RealSense D435i cameras: a fixed head camera and a side camera. Control commands specify 7-DOF joint positions plus a scalar gripper command (8-D action space). Proprio matches the action dimensionality.

\subsection{Real-World Data Collection and Processing}
\label{app:data}

\paragraph{Data collection.}
All compared methods are fine-tuned on the same real-world dataset of 1{,}200 demonstrations, comprising 100 teleoperated episodes for each of the 12 tasks. Demonstrations are collected using a GELLO-based teleoperation system at 20\,FPS. Each timestep contains synchronized RGB observations, robot joint positions, gripper states, and wall-clock timestamps. The timestamps provide the actual interval durations used to compute the numerical motion descriptors. Each episode is paired with a task-specific natural-language instruction, such as ``Pick up the blue ball from the rotating green turntable.''

\paragraph{Training data processing.}
During real-world fine-tuning, each epoch samples 10 windows from every episode without replacement, ensuring balanced contributions across episodes of different lengths. Kinematic descriptors are standardized using the mean and standard deviation computed over all valid motion intervals in the real-world training set.

\subsection{Evaluation Details}
\label{app:eval-protocol}

\paragraph{DOMINO protocol.}
DOMINO evaluation follows its native synchronous protocol. Each 16-step action chunk is executed in full before the next policy query, with background inference and chunk merging disabled. We evaluate 100 episodes per task using the unseen-instruction split and a starting seed of 100{,}000. The control interval is 100\,ms, corresponding to a 1.6\,s chunk horizon. 

\paragraph{Real-world success criterion.}
A trial is considered successful if the robot stably grasps and lifts the moving target within 30\,s without triggering a collision-induced reset.

\paragraph{Visual generalization.}
The evaluated visual shifts modify the background, illumination, or target-object color while preserving the task semantics. The illumination shift is created by reducing the scene brightness relative to the in-distribution setting. OOD-SR is averaged across the three shifts, and retention is defined as $\text{OOD-SR}/\text{ID-SR}$.

\paragraph{Latency measurement.}
Latency is measured over 1{,}000 policy queries on a single NVIDIA GeForce RTX~5090 GPU and includes preprocessing, optical-flow computation, numerical motion-token construction, and full denoising.

\subsection{Baseline Implementation Details}
\label{app:baselines}

All DOMINO baselines are fine-tuned on the same 10{,}500 dynamic demonstrations
using official or author-released training recipes, under a matched data budget
and the benchmark's native synchronous execute-16 protocol. OpenVLA, $\pi_0$,
$\pi_0$-FAST, $\pi_{0.5}$, RDT-1B, InternVLA-M1, InternVLA-A1.5, Isaac-GR00T,
VLA-Adapter, OpenVLA-OFT, and PUMA use checkpoint selection on a held-out
DOMINO validation split. PUMA additionally receives historical optical flow as
specified in its paper. For real-world experiments, InternVLA-A1.5, $\pi_{0.5}$,
and DynamicVLA are initialized from public checkpoints and fine-tuned on our
1{,}200-demonstration dataset with matched augmentation and early stopping on a 10\% validation split.

Ablation variants of DynamicWAM share the same backbone and training schedule.
Naming indicates the removed path:
\begin{itemize}
\item \textbf{w/o flow \& motion}: removes both history-flow frames and kinematic tokens (static WAM).
\item \textbf{w/o motion}: removes kinematic tokens; retains history-flow conditioning.
\item \textbf{w/o flow}: replaces all history-flow frames
with constant black frames, preserving the temporal input
geometry while retaining kinematic-token conditioning.
\item \textbf{w/o RTC}: full model with synchronous chunk execution on the real robot.
\end{itemize}

\paragraph{Sim2Real Performance Gap.}
The substantially different performance of InternVLA-A1.5 in simulation and on the real robot can be primarily attributed to perception--action latency in continuously evolving scenes. Under the native DOMINO evaluation protocol, simulation progression is suspended during policy inference; consequently, the target does not continue moving while an action chunk is being predicted. Its relatively high success rate on DOMINO therefore does not require the policy to compensate for target displacement accumulated during model inference. In the real world, however, the target continues to move throughout the approximately 158ms inference period of InternVLA-A1.5, causing the observation used for prediction to become stale before the generated actions are executed. This perception--execution mismatch is particularly detrimental for tasks requiring accurately timed interception and helps explain its low real-world success rate. DynamicWAM has a comparable per-query latency, but remains effective because its history-flow conditioning explicitly encodes the target's recent motion direction and spatial trajectory. Rather than treating the latest RGB observation as a static snapshot, the model can use this temporal evidence, together with the kinematic tokens, to anticipate the target's near-future position and partially compensate for the displacement occurring during inference. These results indicate that low latency and motion-aware conditioning address complementary aspects of responsive dynamic manipulation.

\section{Architecture and Training Details}
\label{app:architecture-details}

\subsection{DynamicWAM Model Configuration}
\label{app:model-config}

DynamicWAM couples a compact video expert with an action expert through layer-wise joint world--action attention (Figure~\ref{fig:dynamicwam-overview}). Table~\ref{tab:model-config} summarizes the configuration. Observation layouts differ between simulation and real-robot settings: DOMINO uses a head camera with left and right wrist views, while real-robot experiments use two fixed scene cameras.

\begin{table}[!htbp]
\centering
\scriptsize
\begin{tabular}{ll}
\toprule
Component & Configuration \\
\midrule
Teacher video model & Wan2.2-TI2V-5B (Stage~1 only) \\
Compact video expert & 12 layers, $d{=}2048$, FFN 8192, 16 heads \\
Action expert & 12 layers, $d{=}768$, FFN 3072 \\
Action chunk size $H_a$ & 16 \\
Register tokens & 4 \\
History-flow frames $K$ & 4 \\
Kinematic descriptor dim. & 12 per interval (image-plane) \\
Text encoder & UMT5-XXL, max length 512 (frozen) \\
Video VAE & Wan2.2 VAE (frozen) \\
DOMINO views & Head $240{\times}320$ + wrists $120{\times}160$ \\
Real-robot views & Two fixed scene cameras \\
Composite canvas & $384{\times}320$ (center-padded) \\
Future RGB / latent & 8 frames at $192{\times}160$ $\rightarrow$ $[48,2,12,10]$ \\
DOMINO state/action dim. & 14 / 14 \\
Real-robot state/action dim. & 8 / 8 (7 joints + gripper) \\
Stage-3 trainable params & 988.8M (849.2M video + 139.6M action) \\
\bottomrule
\end{tabular}
\caption{DynamicWAM model configuration. The 139.6M action branch includes the kinematic-token module. Future latents are VAE encodings of eight future RGB frames.}
\label{tab:model-config}
\end{table}

\textbf{Token layout.}
At each transformer layer, both video and action queries attend
to the concatenation of video- and action-stream keys and
values, enabling bidirectional joint world--action attention.
3D RoPE is applied to the video tokens. The action stream
contains, in order, one state token, 16 noisy action tokens,
4 kinematic tokens, and 4 register tokens (25 tokens total).
The action decoder reads only the state and action positions,
leaving the robot action representation unchanged. Text
conditions the video stream through cross-attention only and
can influence the action stream indirectly through joint
world--action attention.

\subsection{Optical-Flow Estimation}
\label{app:flow}

\textbf{Flow-matching parameterization.}
Both visual latents and actions are trained with conditional flow matching. Given clean target $x$, noise $\epsilon\sim\mathcal{N}(0,I)$, and noise level $\sigma\in[0,1]$, the corrupted sample is $x^\sigma=(1-\sigma)x+\sigma\epsilon$ and the regression target is $u^\star=\epsilon-x$. Training uses 1{,}000 discrete timesteps with shift 5.0, $\sigma_{\min}=0$, and the extra-one-step discretization.

\subsection{Compact Video Expert and Distillation}
\label{app:distillation}

The compact video expert is distilled from Wan2.2-TI2V-5B.
The 12-layer student is aligned with teacher layers
$\{1,2,4,6,8,11,14,17,20,23,26,30\}$ through
width-preserving projection blocks. Training combines three
complementary objectives.

\paragraph{Ground-truth flow matching.}
We apply the standard conditional flow-matching objective
($\lambda_{\mathrm{gt}}{=}1.0$) to future video latents,
conditioned on the current observation and history-flow frames.

\paragraph{Hidden-state distillation.}
Student layers $\{1,3,5,7,9,12\}$ regress PCA-projected
activations from teacher layers $\{1,4,8,14,20,30\}$.

\paragraph{Motion distillation.}
Student layers $\{6,8,10,12\}$ match the temporal differences
of projected teacher tokens from layers $\{11,17,23,30\}$.

The hidden-state and motion-distillation weights are scheduled
from $0.2$ to $0.1$ and then to $0$ at training progress
$0.35$ and $0.75$, respectively. Timestep sampling combines
uniform, low-, mid-, and high-noise components with
$\sigma$-dependent loss floors. Before Stage~1, PCA statistics
are estimated from 1{,}000 training episodes, using four
subclips per episode and two states per subclip.

\subsection{Multi-Stage Training Procedure}
\label{app:training-stages}

Training comprises three stages, with checkpoint initialization
between Stages~1 and~2.

\paragraph{Stage 1: Video-expert distillation (80{,}000 steps).}
We train the compact WAN and the distillation heads while
keeping the teacher model frozen. The action expert is not
included during this stage.

\paragraph{Checkpoint initialization.}
After Stage~1, we combine the distilled compact WAN with the
action expert initialized from a pretrained DynamicWAM
(w/o motion) checkpoint. The kinematic-token module is
randomly initialized. Action-normalization statistics are
inherited from the same checkpoint and kept fixed during
Stages~2 and~3.

\paragraph{Stage 2: Action-expert pretraining (80{,}000 steps).}
We train the action expert and the kinematic-token module while
keeping the compact WAN frozen. Only the action flow-matching
loss is active ($\lambda_v=0$).

\paragraph{Stage 3: Joint refinement (40{,}000 steps).}
We jointly fine-tune the action expert, the kinematic-token
module, and the compact WAN core while keeping the VAE, text
encoder, and other auxiliary modules frozen. The joint objective
combines the action and video flow-matching losses, with
$\lambda_v$ annealed from 0.01 to 0.001 over the first
2{,}000 steps.

Real-robot fine-tuning follows the same three-stage procedure
with shortened schedules of 8k, 15k, and 8k steps on the
1{,}200-demonstration dataset.

\subsection{Training Hyperparameters}

Table~\ref{tab:hyperparams} summarizes the key training
hyperparameters. Training is conducted on eight NVIDIA H100
GPUs with a per-GPU batch size of 16, yielding a global batch
size of 128. We use AdamW with a weight decay of $10^{-3}$,
gradient clipping at 1.0, bfloat16 precision, and DeepSpeed
ZeRO Stage~0.

\begin{table}[!htbp]
\centering
\small
\begin{tabular}{lccc}
\toprule
Setting & Stage 1 & Stage 2 & Stage 3 \\
\midrule
Steps & 80{,}000 & 80{,}000 & 40{,}000 \\
Video LR & $5{\times}10^{-5}$ & --- & $1{\times}10^{-5}$ \\
Action LR & --- & $5{\times}10^{-5}$ & $5{\times}10^{-5}$ \\
Warmup steps & 1{,}000 & 0 & 0 \\
Min LR ratio & 0.001 & $10^{-5}$ & $10^{-5}$ \\
$\lambda_a$ & --- & 1.0 & 1.0 \\
$\lambda_v$ & 1.0 (GT only) & 0 & $0.01{\to}0.001$ \\
\bottomrule
\end{tabular}
\caption{Training hyperparameters for the DOMINO mainline.}
\label{tab:hyperparams}
\end{table}

\subsection{RTC-Based Asynchronous Execution}
\label{app:rtc}

Real-Time Chunking (RTC) overlaps inference with execution. Let $A_t$ denote the action chunk issued at time $t$ and $H_a$ the chunk horizon. Standard synchronous execution applies $a_t,\ldots,a_{t+H_a-1}$ before querying the policy again, incurring a perception--execution gap equal to inference latency. RTC instead launches inference for $A_{t+H_a}$ while steps $a_t,\ldots,a_{t+H_a-1}$ are still executing. Consecutive chunks are merged with a guidance weight that decays over the overlap region, preventing discontinuities at chunk boundaries.

\textbf{Implementation details.}
On the real robot, RTC maintains a rolling action queue. When remaining queue length falls below a prefetch threshold (4 steps), the policy asynchronously queries the next chunk. Guidance weight $\gamma$ follows the schedule with $\gamma_{\max}{=}0.5$ over a 4-step overlap. Flow history is updated online from the head camera using the same Farneback pipeline as training. DOMINO evaluation intentionally disables RTC to match the benchmark's native synchronous protocol.

\textbf{Inference acceleration.}
Even without RTC, DynamicWAM caches video-branch keys and values after partial video denoising. With 10 action denoising steps, full video forwards run only at steps $\{0,1\}$; subsequent steps reuse cached video activations while updating action tokens only. This reduces per-query latency from $\approx$550\,ms (full joint denoising) to 173.7\,ms without changing the mathematical model.
\section{Motion Representation Construction}
\label{app:motion-construction}

\subsection{Temporal Sampling and Alignment}
\label{app:temporal}

At policy step $t$, we construct $K{=}4$ history intervals with endpoints
$n_k = t-(K-k)\Delta$ for $k=0,\ldots,K$ and $\Delta{=}4$ policy steps.
On the raw converted video, adjacent endpoints are separated by
$\Delta \times r = 12$ frames, where $r{=}3$ is the global temporal downsample rate.
The five endpoints correspond to raw-frame offsets $[48,36,24,12,0]$ relative to $t$.
Interval $k$ spans $[n_{k-1}, n_k]$; intervals whose start lies before episode onset are marked invalid.

Timestamps $\tau_k$ are read from simulator time fields (simulation) or demonstration timestamps (real robot). Interval duration is $\Delta\tau_k=\tau_k-\tau_{k-1}$. Velocities and accelerations are computed in these physical times rather than from container frame rate.

Dense optical flow is computed on the fixed head camera. RGB frames are resized to $64{\times}64$ and converted to grayscale. We use Farneback's algorithm~\cite{farneback2003two} with: pyramid scale 0.5, 3 levels, window size 15, 3 iterations, polynomial expansion $n{=}5$, $\sigma{=}1.2$. The forward flow $\mathbf{U}_k$ maps each pixel in the start frame to its displacement in the end frame.

\subsection{Forward--Backward Consistency Filtering}
\label{app:fb-consistency}

Unreliable flow vectors are removed with a forward--backward consistency test. Given forward flow $\mathbf{U}$ and backward flow $\mathbf{V}$, a pixel $p$ is accepted if the mapped location $p+\mathbf{U}(p)$ lies in-frame and the cycle error
$\lVert \mathbf{U}(p)+\mathbf{V}(p+\mathbf{U}(p))\rVert_2^2$
is below $\epsilon_{\mathrm{rel}}(\lVert\mathbf{U}(p)\rVert_2^2+\lVert\mathbf{V}(\cdot)\rVert_2^2)+\epsilon_{\mathrm{abs}}$,
with $\epsilon_{\mathrm{rel}}{=}0.01$ and $\epsilon_{\mathrm{abs}}{=}0.5$.
If the fraction of accepted pixels falls below 0.1, the entire interval is marked invalid.

\subsection{History-Flow Rendering}
\label{app:rendering}

Accepted flow vectors form the masked field $\hat{\mathbf{U}}_k$. Each field is rendered as an RGB image $R_k=\rho(\hat{\mathbf{U}}_k)$ by mapping direction to hue and magnitude to value:
\begin{equation}
r_k(p)=\min\!\left(\frac{\lVert\hat{\mathbf{U}}_k(p)\rVert_2}{q(\hat{\mathbf{U}}_k)},\,1\right),
\end{equation}
where $q(\cdot)$ is the 99th-percentile magnitude over valid pixels. Per-frame normalization induces the scale invariance.

The history $\mathcal{R}_t=(R_1,\ldots,R_K)$ is concatenated with the current multi-view observation along the temporal axis and encoded by the frozen video VAE.

\subsection{Kinematic Descriptor Construction}
\label{app:kinematic}

For each valid interval $k$ with pixel set $\Omega_k$, we compute
the displacement statistics, interval duration $\Delta\tau_k$,
velocity statistics, and acceleration statistics, yielding a
12-dimensional numerical motion descriptor. Table~\ref{tab:kinematic-statistics}
reports the dataset-level mean and standard deviation of each
descriptor dimension, computed over 1.8 million valid intervals
in the DOMINO training set. These statistics are used to standardize
the numerical motion descriptors before they are provided to the
action expert.

For invalid intervals, all stored feature values are set to zero
and the interval-validity flag is set to false. When a valid
predecessor is unavailable, the acceleration vector and its
magnitude are masked and replaced with a learned
invalid-acceleration embedding during inference.

\begin{table}[!htbp]
\centering
\small
\setlength{\tabcolsep}{3pt}
\begin{tabular}{lrr}
\toprule
Feature & Mean & Std \\
\midrule
Mean $\Delta x$ (px) & 0.020 & 0.531 \\
Mean $\Delta y$ (px) & 0.122 & 0.874 \\
Mean $\lVert\Delta\rVert$ (px) & 0.896 & 0.784 \\
$p_{99}$ $\lVert\Delta\rVert$ (px) & 4.517 & 2.938 \\
$\Delta\tau$ (s) & 0.645 & 0.070 \\
Mean $v_x$ (px/s) & 0.031 & 0.848 \\
Mean $v_y$ (px/s) & 0.206 & 1.450 \\
Mean speed (px/s) & 1.434 & 1.313 \\
$p_{99}$ speed (px/s) & 7.091 & 4.706 \\
Mean $a_x$ (px/s$^2$) & $-0.007$ & 1.659 \\
Mean $a_y$ (px/s$^2$) & $-0.027$ & 2.805 \\
Mean $\lVert a\rVert$ (px/s$^2$) & 2.356 & 2.251 \\
\bottomrule
\end{tabular}
\caption{DOMINO training-set kinematic statistics. Units are flow-grid pixels and simulator seconds.}
\label{tab:kinematic-statistics}
\end{table}

\section{Extended Experimental Results}
\label{app:extended-results}

\subsection{Full Per-Task Results on DOMINO}
\label{app:domino-per-task}

Table~\ref{tab:domino-per-task} reports detailed per-task success rates
and manipulation scores for the four DynamicWAM variants on DOMINO
Level~1. Each variant is evaluated on 100 episodes per task, resulting
in 3{,}500 evaluation episodes per variant.

\begin{table*}[!tbp]
\centering
\normalsize
\setlength{\tabcolsep}{1.5pt}
\renewcommand{\arraystretch}{1.10}
\begin{tabular}{lrrrrrrrr}
\toprule
&
\multicolumn{8}{c}{DynamicWAM Variants} \\
\cmidrule(lr){2-9}

Task
& \multicolumn{2}{c}{
\begin{tabular}[c]{@{}c@{}}
w/o Flow\\
\& Motion
\end{tabular}}
& \multicolumn{2}{c}{
\begin{tabular}[c]{@{}c@{}}
w/o\\
Motion
\end{tabular}}
& \multicolumn{2}{c}{
\begin{tabular}[c]{@{}c@{}}
w/o\\
Flow
\end{tabular}}
& \multicolumn{2}{c}{Full} \\
\cmidrule(lr){2-3}
\cmidrule(lr){4-5}
\cmidrule(lr){6-7}
\cmidrule(lr){8-9}

& SR$\uparrow$ & MS$\uparrow$
& SR$\uparrow$ & MS$\uparrow$
& SR$\uparrow$ & MS$\uparrow$
& SR$\uparrow$ & MS$\uparrow$ \\
\midrule

\textit{Move Pillbottle Pad}
& 42 & 48.1 & 55 & 59.4 & 65 & 70.1 & 91 & 91.4 \\

\textit{Place Bread Basket}
& 27 & 35.5 & 46 & 50.5 & 52 & 57.9 & 87 & 88.5 \\

\textit{Beat Block Hammer}
& 36 & 45.2 & 49 & 55.4 & 46 & 53.1 & 84 & 86.7 \\

\textit{Place Shoe}
& 69 & 74.2 & 90 & 92.0 & 91 & 92.5 & 91 & 92.8 \\

\textit{Place Empty Cup}
& 39 & 44.8 & 60 & 63.4 & 54 & 59.8 & 87 & 88.5 \\

\textit{Place Phone Stand}
& 33 & 44.0 & 32 & 43.7 & 44 & 54.1 & 80 & 84.6 \\

\textit{Place Container Plate}
& 47 & 51.6 & 46 & 49.9 & 62 & 64.2 & 77 & 79.0 \\

\textit{Stamp Seal}
& 35 & 42.7 & 49 & 58.1 & 56 & 62.5 & 68 & 73.1 \\

\textit{Place Fan}
& 20 & 25.3 & 32 & 35.2 & 43 & 46.8 & 67 & 69.9 \\

\textit{Place Rat Pad}
& 24 & 33.5 & 25 & 37.8 & 37 & 46.8 & 54 & 61.3 \\

\textit{Press Stapler}
& 48 & 52.5 & 38 & 43.8 & 25 & 34.0 & 53 & 60.2 \\

\textit{Place Object Stand}
& 33 & 38.5 & 32 & 36.2 & 40 & 44.2 & 55 & 58.4 \\

\textit{Grab Roller}
& 27 & 51.4 & 34 & 58.3 & 33 & 60.7 & 45 & 68.2 \\

\textit{Shake Bottle Horiz.}
& 23 & 39.7 & 36 & 53.7 & 47 & 65.9 & 43 & 62.2 \\

\textit{Move Stapler Pad}
& 16 & 28.5 & 12 & 24.2 & 22 & 33.5 & 39 & 47.1 \\

\textit{Place Object Scale}
& 12 & 21.2 & 24 & 31.3 & 24 & 32.6 & 36 & 42.6 \\

\textit{Shake Bottle}
& 24 & 36.8 & 33 & 47.5 & 44 & 58.6 & 33 & 49.2 \\

\textit{Dump Bin Bigbin}
& 11 & 15.8 & 17 & 23.0 & 25 & 30.4 & 30 & 34.9 \\

\textit{Hanging Mug}
& 11 & 36.1 & 23 & 38.5 & 26 & 44.0 & 26 & 40.2 \\

\textit{Put Bottles Dustbin}
& 33 & 49.8 & 31 & 42.2 & 28 & 42.9 & 24 & 41.2 \\

\textit{Click Alarmclock}
& 11 & 14.4 & 16 & 18.8 & 20 & 23.0 & 23 & 25.4 \\

\textit{Handover Mic}
& 20 & 33.4 & 28 & 43.4 & 13 & 33.5 & 22 & 39.7 \\

\textit{Place Object Basket}
& 28 & 48.8 & 25 & 39.4 & 27 & 50.8 & 22 & 42.1 \\

\textit{Handover Block}
& 1 & 40.7 & 8 & 44.3 & 11 & 59.0 & 15 & 48.8 \\

\textit{Adjust Bottle}
& 16 & 46.0 & 19 & 45.5 & 16 & 50.0 & 14 & 50.3 \\

\textit{Move Playingcard Away}
& 8 & 20.8 & 9 & 19.3 & 31 & 51.2 & 13 & 38.2 \\

\textit{Move Can Pot}
& 14 & 49.4 & 7 & 47.2 & 17 & 55.4 & 11 & 46.4 \\

\textit{Rotate Qrcode}
& 15 & 23.6 & 9 & 18.5 & 15 & 25.0 & 11 & 21.2 \\

\textit{Place Bread Skillet}
& 9 & 39.0 & 14 & 42.7 & 12 & 39.3 & 9 & 35.3 \\

\textit{Place A2B Left}
& 6 & 32.2 & 9 & 31.3 & 5 & 31.1 & 8 & 31.9 \\

\textit{Put Object Cabinet}
& 12 & 41.7 & 17 & 50.0 & 6 & 48.3 & 7 & 46.5 \\

\textit{Place A2B Right}
& 7 & 35.0 & 12 & 35.1 & 6 & 33.0 & 6 & 33.6 \\

\textit{Place Can Basket}
& 21 & 46.2 & 13 & 36.2 & 9 & 44.8 & 5 & 39.8 \\

\textit{Scan Object}
& 8 & 38.4 & 1 & 31.9 & 13 & 37.8 & 1 & 29.7 \\

\textit{Click Bell}
& 9 & 16.2 & 2 & 9.3 & 2 & 13.1 & 0 & 11.7 \\

\midrule
\textbf{Overall}
& 22.7 & 38.3
& 27.2 & 41.6
& 30.5 & 47.2
& \textbf{38.2} & \textbf{53.2} \\
\bottomrule
\end{tabular}

\caption{
Detailed per-task results of DynamicWAM variants on DOMINO Level~1.
Each variant is evaluated on 100 episodes per task. SR and MS denote
success rate (\%) and manipulation score, respectively. The best
overall results are shown in bold. ``w/o Flow'' is the matched
black-flow information-null control, which retains the motion-token
path and input geometry.
}
\label{tab:domino-per-task}
\end{table*}

\subsection{Full Per-Task Real-World Results}
\label{app:realworld-per-task}

Table~\ref{tab:realworld-per-task} reports detailed per-task success
rates for the external baselines and DynamicWAM variants. We additionally
report the average performance within each motion level and across all
12 tasks.

\begin{table*}[!tbp]
\centering
\small
\setlength{\tabcolsep}{4.0pt}
\renewcommand{\arraystretch}{1.05}
\begin{tabular}{llrrrrrr}
\toprule
Level
& Task
& \multicolumn{2}{c}{External Baselines}
& \multicolumn{4}{c}{DynamicWAM Variants} \\
\cmidrule(lr){3-4}
\cmidrule(lr){5-8}
&
& $\pi_{0.5}$
& DynamicVLA
& \begin{tabular}[c]{@{}c@{}}
w/o Flow\\
\& Motion
\end{tabular}
& \begin{tabular}[c]{@{}c@{}}
w/o\\
Motion
\end{tabular}
& \begin{tabular}[c]{@{}c@{}}
w/o\\
Flow
\end{tabular}
& \begin{tabular}[c]{@{}c@{}}
Full\\
(RTC)
\end{tabular} \\
&
& SR$\uparrow$
& SR$\uparrow$
& SR$\uparrow$
& SR$\uparrow$
& SR$\uparrow$
& SR$\uparrow$ \\
\midrule

\textbf{L1}
& \textit{Line Car}
& 25.00 & 30.00 & 35.00 & 50.00 & 50.00 & \textbf{70.00} \\
&
\textit{Line Tennis}
& 45.00 & 40.00 & 45.00 & 55.00 & 60.00 & \textbf{75.00} \\
&
\textit{Line Cube}
& 20.00 & 15.00 & 25.00 & 40.00 & 45.00 & \textbf{55.00} \\
&
\textit{Line Rat}
& 40.00 & 35.00 & 45.00 & 60.00 & 65.00 & \textbf{80.00} \\
\cmidrule(lr){2-8}
&
\textit{Level Average}
& 32.50 & 30.00 & 37.50 & 51.25 & 55.00 & \textbf{70.00} \\
\midrule

\textbf{L2}
& \textit{TT-BlueBall-Yellow}
& 45.00 & 30.00 & 45.00 & 50.00
& \textbf{55.00} & \textbf{55.00} \\
&
\textit{TT-Rat-Yellow}
& 30.00 & 35.00 & 30.00 & 45.00 & 45.00 & \textbf{50.00} \\
&
\textit{TT-Cube-Yellow}
& 25.00 & 20.00 & 25.00 & 30.00 & 35.00 & \textbf{40.00} \\
&
\textit{TT-Snack-Yellow}
& 55.00 & 50.00 & 40.00 & 45.00 & 55.00 & \textbf{60.00} \\
\cmidrule(lr){2-8}
&
\textit{Level Average}
& 38.75 & 33.75 & 35.00 & 42.50 & 47.50 & \textbf{51.25} \\
\midrule

\textbf{L3}
& \textit{Complex Tennis}
& 0.00 & 0.00 & 5.00 & 20.00
& \textbf{25.00} & \textbf{25.00} \\
&
\textit{Complex Cube A}
& 0.00 & 0.00 & 0.00 & 5.00
& \textbf{10.00} & \textbf{10.00} \\
&
\textit{Complex Cube B}
& 0.00 & 0.00 & 0.00 & 10.00
& \textbf{15.00} & \textbf{15.00} \\
&
\textit{Complex Rat}
& 0.00 & 0.00 & 5.00
& \textbf{25.00} & 20.00 & \textbf{25.00} \\
\cmidrule(lr){2-8}
&
\textit{Level Average}
& 0.00 & 0.00 & 2.50 & 15.00 & 17.50 & \textbf{18.75} \\
\midrule

&
\textbf{Overall Average}
& 23.75 & 21.25 & 25.00 & 36.25 & 40.00 & \textbf{46.67} \\
\bottomrule
\end{tabular}

\caption{
Detailed per-task real-world success rates (\%). L1--L3 denote linear,
circular, and compound target motion, respectively. Each task is
evaluated over 20 real-robot trials. The best result in each row is
shown in bold; all tied best results are bolded.
}
\label{tab:realworld-per-task}
\end{table*}

\subsection{Visual Generalization under Appearance Shifts}
\label{app:visual-generalization}

We further evaluate the visual generalization of DynamicWAM on all four
Level~1 tasks---Line Car, Line Tennis, Line Cube, and Line Rat---under
three appearance shifts: background, object-color, and lighting shifts.
In each condition, only the visual appearance of the scene is modified,
while the robot configuration, camera viewpoints, target trajectory,
language instruction, and success criterion remain unchanged.

For each task and visual condition, we conduct 20 real-robot trials.
The in-distribution condition uses the same visual configuration as the
training demonstrations. The background condition changes the surrounding
scene appearance, while the object-color condition changes the target color
to a value unseen during training. The lighting condition reduces the
overall scene illumination. All other evaluation settings, including
target motion, robot initialization, control frequency, and trial duration,
remain unchanged.

% \begin{figure*}[t]
%     \centering
%     \includegraphics[
%         width=0.66\textwidth
%     ]{Figures/scalability.pdf}
%     \caption{
%     Representative visual conditions for the TT-Snack-Yellow task
%     in the real-world visual-generalization evaluation.
%     Relative to the original in-distribution setting in (a), we modify
%     only the turntable color in (b), the target-object color in (c), or
%     the scene illumination in (d). The robot configuration, camera
%     viewpoints, target trajectory, task instruction, and success
%     criterion remain unchanged across conditions.
%     }
%     \label{fig:visual-generalization}
% \end{figure*}

Table~\ref{tab:visual-generalization-full} reports the aggregate
results. DynamicWAM achieves the highest success rate under all three
appearance shifts, reaching an average out-of-distribution success rate
of 36.3\%, compared with 17.5\% for $\pi_{0.5}$ and 20.0\% for the
current-frame WAM. The largest absolute advantage is observed under
lighting and object-color shifts. Background changes remain the most
challenging condition, as they modify a large visual region surrounding
the moving target.

% \begin{table}[!htbp]
% \centering
% \begingroup
% \small
% \setlength{\tabcolsep}{1.5pt}
% \begin{tabular}{lcccccc}
% \toprule
% Method
% & ID
% & Turntable
% & Object
% & Lighting
% & OOD Avg.
% & Retain \\
% \midrule
% $\pi_{0.5}$
% & 32.5
% & 12.5
% & 22.5
% & 17.5
% & 17.5
% & 0.54 \\

% \begin{tabular}[c]{@{}l@{}}
% DynamicWAM\\
% (current frame)
% \end{tabular}
% & 37.5
% & 6.3
% & 23.8
% & 30.0
% & 20.0
% & 0.53 \\

% DynamicWAM (ours)
% & \textbf{70.0}
% & \textbf{25.0}
% & \textbf{40.0}
% & \textbf{43.8}
% & \textbf{36.3}
% & 0.52 \\
% \bottomrule
% \end{tabular}
% \endgroup
% \caption{
% Aggregate success rates (\%) under visual appearance shifts.
% OOD Avg. averages the turntable-color, object-color, and lighting
% conditions. Retain denotes OOD Avg./ID.
% }
% \label{tab:visual-generalization-summary}
% \end{table}

% Table~\ref{tab:visual-generalization-full} provides the complete
% task-level results. DynamicWAM consistently outperforms the evaluated
% baselines across the majority of tasks and visual conditions. These
% results indicate that the benefit of motion conditioning persists when
% the appearance of either the moving target or its surrounding context
% differs from the training distribution.

\begin{table*}[!tbp]
\centering
\small
\setlength{\tabcolsep}{3.5pt}
\renewcommand{\arraystretch}{1.05}
\begin{tabular}{clccccc}
\toprule
Method
& Task
& ID
& Background
& \begin{tabular}[c]{@{}c@{}}Object\\Color\end{tabular}
& Lighting
& OOD Avg. \\
\midrule

\multirow{5}{*}{$\pi_{0.5}$}
& Line Car
& 25.0 & 20.0 & 35.0 & 25.0 & 26.7 \\

& Line Tennis
& 45.0 & 5.0 & 10.0 & 10.0 & 8.3 \\

& Line Cube
& 20.0 & 10.0 & 20.0 & 15.0 & 15.0 \\

& Line Rat
& 40.0 & 15.0 & 25.0 & 20.0 & 20.0 \\

\cmidrule(lr){2-7}
& \textit{Average}
& 32.5 & 12.5 & 22.5 & 17.5 & 17.5 \\

\midrule

\multirow{5}{*}{
  \shortstack[c]{DynamicWAM\\(w/o flow \& motion)}
}
& Line Car
& 35.0 & 10.0 & 35.0 & 40.0 & 28.3 \\

& Line Tennis
& 45.0 & 5.0 & 20.0 & 20.0 & 15.0 \\

& Line Cube
& 25.0 & 5.0 & 15.0 & 30.0 & 16.7 \\

& Line Rat
& 45.0 & 5.0 & 25.0 & 30.0 & 20.0 \\

\cmidrule(lr){2-7}
& \textit{Average}
& 37.5 & 6.3 & 23.8 & 30.0 & 20.0 \\

\midrule

\multirow{5}{*}{
  \shortstack[c]{DynamicWAM\\(ours)}
}
& Line Car
& 70.0 & 35.0 & 60.0 & 55.0 & 50.0 \\

& Line Tennis
& 75.0 & 15.0 & 25.0 & 30.0 & 23.3 \\

& Line Cube
& 55.0 & 20.0 & 20.0 & 40.0 & 26.7 \\

& Line Rat
& 80.0 & 30.0 & 55.0 & 50.0 & 45.0 \\

\cmidrule(lr){2-7}
& \textit{Average}
& \textbf{70.0}
& \textbf{25.0}
& \textbf{40.0}
& \textbf{43.8}
& \textbf{36.3} \\

\bottomrule
\end{tabular}

\caption{
Complete task-level success rates (\%) on the four Level~1 tasks under
visual appearance shifts. Each entry is evaluated over 20 real-robot
trials. OOD Avg. averages the background, object-color, and lighting
conditions.
}
\label{tab:visual-generalization-full}
\end{table*}

\subsection{Effect of Real-Time Execution Components}
\label{app:execution-strategies}

We further disentangle the contributions of the real-time execution
infrastructure and RTC using three configurations of the full DynamicWAM
model. The standard synchronous configuration disables both our real-time
execution core (RealCore) and RTC. The RealCore-only configuration enables
the real-time execution core while keeping RTC disabled. The full
configuration additionally enables RTC for asynchronous action-chunk
generation and execution. All three configurations use identical model
weights, observation histories, action-chunk horizons, and inference
settings, differing only in their execution mechanisms.

\begin{table}[!htbp]
\centering
\begingroup
\small
\setlength{\tabcolsep}{5pt}
\begin{tabular}{lccc}
\toprule
Execution Configuration & RealCore & RTC & Avg. SR$\uparrow$ \\
\midrule
Standard synchronous
& No & No & 37.50 \\

RealCore only
& Yes & No & 42.08 \\

Full RTC
& Yes & Yes & \textbf{46.67} \\
\bottomrule
\end{tabular}
\endgroup
\caption{
Ablation of real-time execution components on the 12-task
real-world evaluation. All configurations use the full DynamicWAM
model and differ only in whether RealCore and RTC are enabled.
Results report average success rates (\%).
}
\label{tab:execution-component-ablation}
\end{table}

As shown in Table~\ref{tab:execution-component-ablation}, standard
synchronous execution achieves an average success rate of 37.50\%.
Enabling RealCore while keeping RTC disabled improves performance to
42.08\%, a gain of 4.58 percentage points. Further enabling RTC raises
the average success rate to 46.67\%, providing an additional gain of
4.59 points over the RealCore-only configuration and an overall gain
of 9.17 points over standard synchronous execution. These results
separate the benefit of the real-time execution core from the
additional improvement introduced by RTC-based asynchronous
action-chunk execution.

\subsection{Future-Frame Reconstruction Quality}
\label{app:reconstruction}

We evaluate Stage-1 future-video generation on 500 held-out DOMINO packed
windows. The compact video expert samples future latents with
10 flow-matching steps, conditioned on the packed condition observation
latents and the language embedding. We report latent RMSE against
ground-truth future latents, and pixel RMSE/PSNR after decoding both
predicted and ground-truth future latents with the frozen Wan2.2 VAE under a
shared first-latent protocol; pixel metrics are computed on RGB frames in
$[0,1]$ at $192{\times}160$.

\begin{figure*}[t]
\centering
\includegraphics[width=0.88\textwidth]{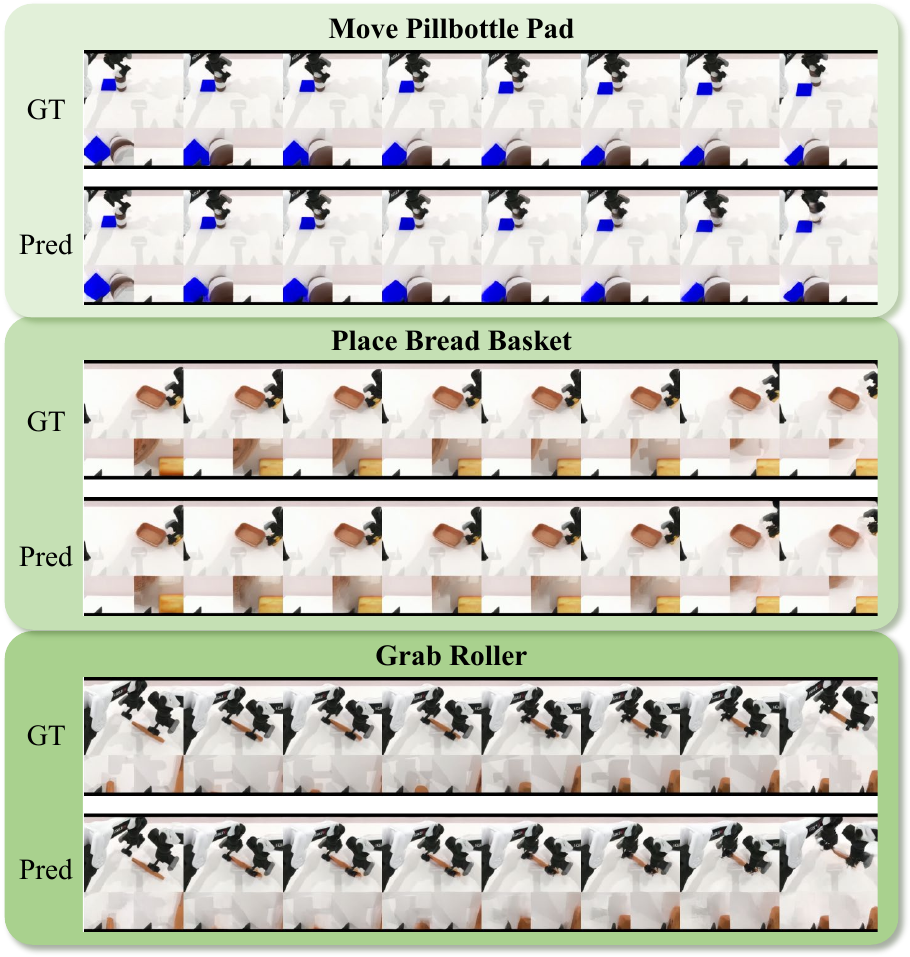}
\caption{
Stage-1 future-frame reconstruction on three DOMINO tasks.
For each task, the top row shows ground-truth future frames and the bottom
row shows Stage-1 predictions (eight frames at $192{\times}160$).
}
\label{fig:stage1-recon}
\end{figure*}

The Stage-1 expert achieves a latent RMSE of 0.291, a pixel RMSE of 0.0565,
and a PSNR of 25.0\,dB. Figure~\ref{fig:stage1-recon} shows qualitative
GT/Pred comparisons on three representative tasks---Move Pillbottle Pad,
Place Bread Basket, and Grab Roller---decoded with the real condition frame
as the first-latent anchor. Predictions preserve object identity, gripper--object
spatial relations, and short-horizon motion; residual discrepancies appear
mainly as mild edge softening on fast-moving regions in later frames.

\subsection{Motion-Conditioning Attention Visualization}
\label{app:motion-attention}

To illustrate the two motion cues used by Stage-3 DynamicWAM, we visualize
(i)~the history-flow conditioning signal and (ii)~kinematic-token attention
on held-out DOMINO packed windows. History flow is shown as a magnitude
overlay on the head-camera RGB. For kinematic attention, because
FlashAttention does not return weights, we recompute
$\mathrm{softmax}(Q_{\mathrm{motion}}K_{\mathrm{condition}}^{\top}/\sqrt{d})$
from the same projected queries and keys used in joint world--action
attention, with kinematic motion tokens as queries and condition video
tokens as keys. Maps are averaged over attention heads and the four motion
tokens, then over late joint-attention layers (indices 8, 10, and 11) at a
fixed flow-matching noise level $\sigma{=}0.15$. The resulting spatial map
lives on the condition latent grid and is projected onto the head-camera
RGB for display.

\begin{figure*}[t]
\centering
\includegraphics[width=0.88\textwidth]{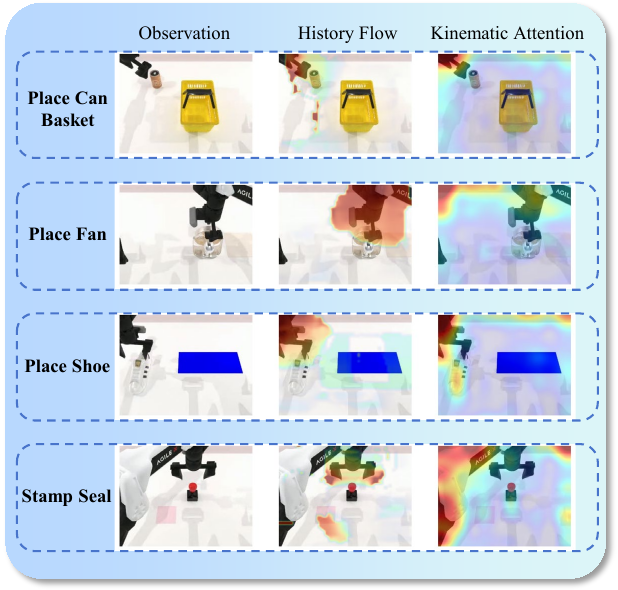}
\caption{
Motion-conditioning visualization on Stage-3 DynamicWAM.
For each task: observation (left), history-flow magnitude overlay (middle),
and kinematic$\to$condition attention on the head view (right).
}
\label{fig:motion-attention}
\end{figure*}

Figure~\ref{fig:motion-attention} shows four representative tasks.
History-flow overlays concentrate energy on recently moving regions
(gripper and manipulated object), providing a spatially localized motion
cue. Kinematic-to-image attention is coarser---each latent cell covers a
large image region---so the upsampled heatmap is diffuse rather than
pixel-accurate, while typically peaking near the active manipulator region.
Together with Figure 4 , these visualizations are
consistent with a complementary role of the two pathways: flow supplies
local motion structure, whereas kinematic tokens provide a coarser
attentional bias over the observation latent grid.

\subsection{Additional Ablation Studies}
\label{app:additional-ablations}
\paragraph{Representative per-task ablation.}
Table~\ref{tab:ablation-per-task} reports selected tasks discussed in Section 4.4.

\begin{table}[!htbp]
\centering
\small
\setlength{\tabcolsep}{3pt}
\begin{tabular}{lcccc}
\toprule
Task & \shortstack{w/o Flow\\\& Motion} & w/o Motion & w/o Flow & Full \\
\midrule
Move Pillbottle Pad & 42 & 55 & 65 & \textbf{91} \\
Beat Block Hammer & 36 & 49 & 46 & \textbf{84} \\
Place Phone Stand & 33 & 32 & 44 & \textbf{80} \\
Put Object Cabinet & 12 & \textbf{17} & 6 & 7 \\
Place Can Basket & \textbf{21} & 13 & 9 & 5 \\
\bottomrule
\end{tabular}
\caption{Per-task ablation success rates (\%) on DOMINO.}
\label{tab:ablation-per-task}
\end{table}

\paragraph{History length and interval stride.}
Using $K{=}2$ history intervals reduces SR to 31.4\%; $K{=}8$ improves SR marginally to 38.6\% at 1.4$\times$ latency. Policy stride $\Delta{=}4$ is optimal; $\Delta{=}2$ is too myopic (34.1\% SR) and $\Delta{=}8$ overly smooths motion (35.0\% SR).

\paragraph{Flow rendering percentile.}
Replacing $p_{99}$ normalization with $p_{95}$ or per-dataset fixed scaling reduces SR by 1.2 and 2.8 points respectively, confirming that per-frame normalization is useful for spatial structure but insufficient without kinematic tokens.

\subsection{Failure Case Analysis}
\label{app:failures}

We categorize failure modes across 50 unsuccessful real-robot trials:

\begin{itemize}
\item \textbf{Timing error (42\%):} gripper closes early or late relative to target passage, most common on Level~1 linear tasks at higher speeds.
\item \textbf{Spatial misalignment (26\%):} approach direction correct but lateral offset exceeds grasp tolerance; often preceded by stale flow history after occlusions.
\item \textbf{Tracking loss (18\%):} target leaves the field of view or flow consistency falls below threshold, yielding invalid kinematic tokens.
\item \textbf{Slip after contact (10\%):} partial grasps on smooth or deformable objects (ZP Mimi Yellow).
\item \textbf{Other (4\%):} safety stops, communication timeouts.
\end{itemize}

RTC reduces timing-error failures by 34\% relative to synchronous execution by shrinking the inference--execution gap. Kinematic tokens provide the largest reduction in early-contact failures on Level~2 circular tasks, where constant speed is ambiguous from flow direction alone.

On DOMINO, common failures include missed interception on fast linear tasks, incorrect contact orientation on Scan Object and Click Bell, and long-horizon drift on multi-step tasks such as Put Object Cabinet. History-flow conditioning primarily reduces interception misses; kinematic tokens help tasks requiring speed-dependent contact timing (Place Phone Stand).
\end{document}